\documentclass[journal]{IEEEtran}

\usepackage[utf8]{inputenc}
\usepackage{hyperref}

\usepackage{graphicx}
\usepackage{subfigure}
\usepackage{multirow}

\usepackage{amsmath}
\usepackage{amsfonts}
\usepackage{algorithm}
\usepackage{algpseudocode}

\usepackage{color}
\usepackage{xspace}

\newcommand*{\eg}{{\em e.g.}\xspace}
\newcommand*{\ie}{{\em i.e.}\xspace}
\newcommand*{\etc}{{\em etc}\xspace}
\newcommand*{\etal}{{\em et al.}\xspace}
\newcommand*{\wrt}{w.r.t.\xspace}
\usepackage{amssymb}
\usepackage{pifont}
\newcommand{\xmark}{\ding{55}}%

\def\SexyName{SCGN}

\begin{document}
	
	
	\title{Deep View Synthesis via Self-Consistent \\
		Generative Network}
	
	\author{Zhuoman~Liu$^*$,~Wei~Jia$^*$,~Ming~Yang,~Peiyao~Luo,~Yong~Guo,~and~Mingkui~Tan$^\dag$
		
		\thanks{Z. Liu, P. Luo, Y. Guo, and M. Tan are with School of Software Engineering, South China University of Technology, Guangzhou 510640, China. Z. Liu is also with Pazhou Laboratory, Guangzhou 510335, China. E-mail: selzm@mail.scut.edu.cn \{is.luopeiyao, guoyongcs\}@gmail.com mingkuitan@scut.edu.cn.}
		\thanks{Z. Liu, W. Jia, and M. Yang are with CVTE Research, Guangzhou 510530, China. E-mail: \{liuzhuoman, jiawei, yangming\}@cvte.com.}
		\thanks{This paper has supplementary downloadable material available at http://ieeexplore.ieee.org., provided by the author. The material includes more experimental results. This material is 1.83MB in size.}
		\thanks{$*$ Authors contributed equally. $^\dag$ Corresponding author.}}

	\markboth{Journal of \LaTeX\ Class Files,~Vol.~14, No.~26, January~2021}%
	{Shell \MakeLowercase{\textit{et al.}}: Bare Demo of IEEEtran.cls for IEEE Journals}
	
	\maketitle
	\begin{abstract}
		View synthesis aims to produce unseen views from a set of views captured by two or more cameras at different positions. 
		This task is non-trivial since it is hard to conduct pixel-level matching among different views.
		To address this issue, most existing methods seek to exploit the geometric information to match pixels.
		However, when the distinct cameras have a large baseline (\ie, far away from each other), severe geometry distortion issues would occur and the geometric information may fail to provide useful guidance,
		resulting in very blurry synthesized images. 
		To address the above issues,
		in this paper, we propose a novel deep generative model, called Self-Consistent Generative Network (\SexyName), which synthesizes novel views from the given input views without explicitly exploiting the geometric information.
		The proposed \SexyName\ model consists of two main components, \ie, a View Synthesis Network (VSN) and a View Decomposition Network (VDN), both employing an Encoder-Decoder structure. 
		Here, the VDN seeks to reconstruct input views from the synthesized novel view to preserve the consistency of view synthesis.
		Thanks to VDN, \SexyName\ is able to synthesize novel views without using any geometric rectification before encoding, making it easier for both training and applications. 
		Finally, adversarial loss is introduced to improve the photo-realism of novel views. 
		Both qualitative and quantitative comparisons against several state-of-the-art methods on two benchmark tasks demonstrated the superiority of our approach.
	\end{abstract}
	
	
	\IEEEpeerreviewmaketitle
	
	
	\section{Introduction}
\begin{figure}
		\centering
		\vspace{-0.15in}
		\includegraphics[width=0.4\textwidth]{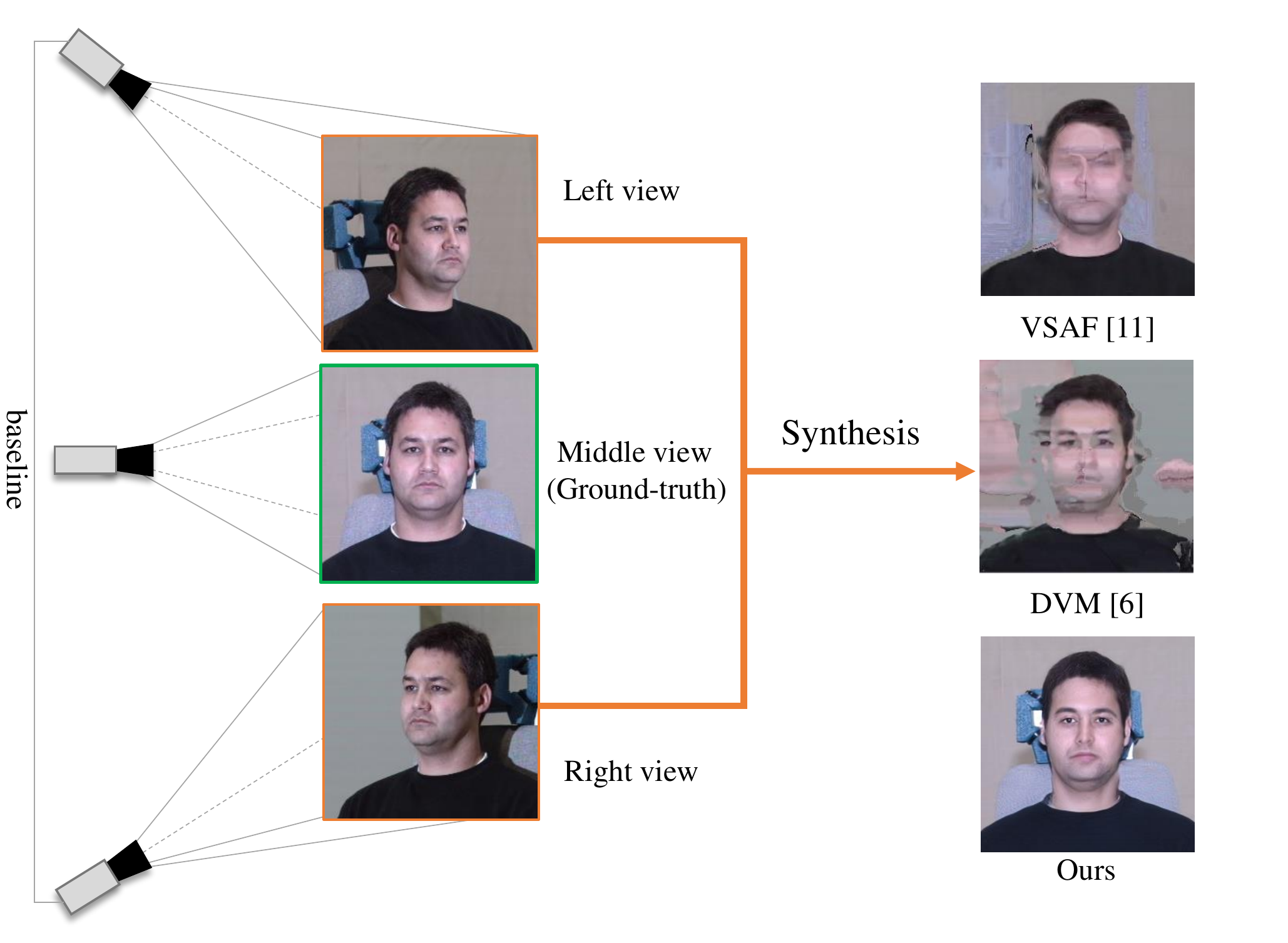}
		\vspace{-0.1in}
		\caption{
			A view synthesis example: middle view (frontal) synthesis from the left and right views with a large baseline (namely, a large distance between two camera views) on the Multi-PIE~\cite{GROSS2010807} dataset. 
			The left and right views are captured by two distinct cameras and then used to synthesize the missing middle view. 
			This task can be applied in a real industrial scenario --- a real-time video conferencing system developed by the company with which the authors are working. 
		}
		\vspace{-0.2in}
		\label{fig:comp_intro}
	\end{figure}

	\IEEEPARstart{V}{iew} synthesis generates a novel (absent) camera view image  from known camera views of the same scene, as shown in Fig. \ref{fig:comp_intro}. 
	It can be widely applied in video conferencing~\cite{atzpadin2004stereo}, virtual reality~\cite{scharstein1996stereo}, and free-viewpoint TV~\cite{oh2009hole}, \etc. 
	In this paper, we focus on synthesizing a middle view from two different views in real industrial scenarios where the ideal view is hard to obtain due to hardware limitations.
	For example, in some video conferencing equipment, cameras are positioned symmetrically on each side of the screen with a large baseline (\ie, the distance between two camera views~\cite{zhou2018stereo}).
	Moreover, the baseline between the two cameras is often pre-defined for specific products.
	
	The view synthesis task, however, is extremely difficult due to the following challenges: 1) The large distance between two camera views may lead to huge occlusion. The model is hard to synthesize the novel views given limited information. 2) View synthesis is an ill-posed problem. Specifically, there exists an infinite number of middle/novel views that correspond to the same input views \cite{huang2017}. Thus, the space of the possible view synthesis functions can be extremely large, making it hard to find a good solution.
	Several recent works~\cite{dibr2018, array2018, NIPS2019_8977} attempt to solve the view synthesis problem by warping with a depth camera. However, depth images are difficult to obtain due to the limitation of the hardware. Therefore, some geometry-based view synthesis methods~\cite{tatarchenko2016multi, zhou2016view, Flynn_2019_CVPR, Ding2017,sun2018multi, zhou2018stereo} are proposed to synthesize novel views without the depth image. Such approaches add geometry constraints to preserve consistency between input views and the synthesized view. However, when the input views have huge occlusion, these geometry-based methods may learn mismatching corresponding map between the input views, or even fail to learn the correspondences. To overcome the drawbacks of geometry-based methods, some image-content-based methods~\cite{dosovitskiy2015learning},~\cite{tatarchenko2016multi},~\cite{zhang2019structure} formulate the view synthesis task as a mapping from input views to the target view without geometry constraints. Despite these attempts, the space of the possible view synthesis functions is still extremely large, which makes it difficult to learn a good model.

    To address the above issues, we propose a new view synthesis method, called Self-Consistent Generative Network (\SexyName), to simultaneously produce photo-realistic novel views and preserve consistency among different views of the same scene. 
    Specifically, to address the challenge brought by the large baseline between cameras, we propose View Synthesis Network (VSN) model that directly learns the mapping from the input side views to the resultant novel/middle view.
	To reduce the space of possible mapping functions, we design a self-consistency scheme that introduces an additional constraint by decomposing the synthesized novel view back into the original input side views, and propose a View Decomposition Network (VDN) to learn the decomposition mapping.
    To further improve the photo-realism of the synthesized view, we incorporate an adversarial loss and an image sharpness loss into the training objective to train the proposed model.
     Unlike cycle-consistency~\cite{zhu2017unpaired} that helps minimize the distribution divergence, our self-consistency builds a cycle to improve the pixel-wise prediction. With the self-consistency constraint, we are able to effectively reduce the space of possible mapping functions and thus obtain promising views synthesis performance.
     Extensive experiments on an indoor dataset and an outdoor dataset demonstrate the superiority of the proposed method over existing methods.

	\begin{figure*}[!htp]
		\begin{center}
			\includegraphics[width=0.9\textwidth]{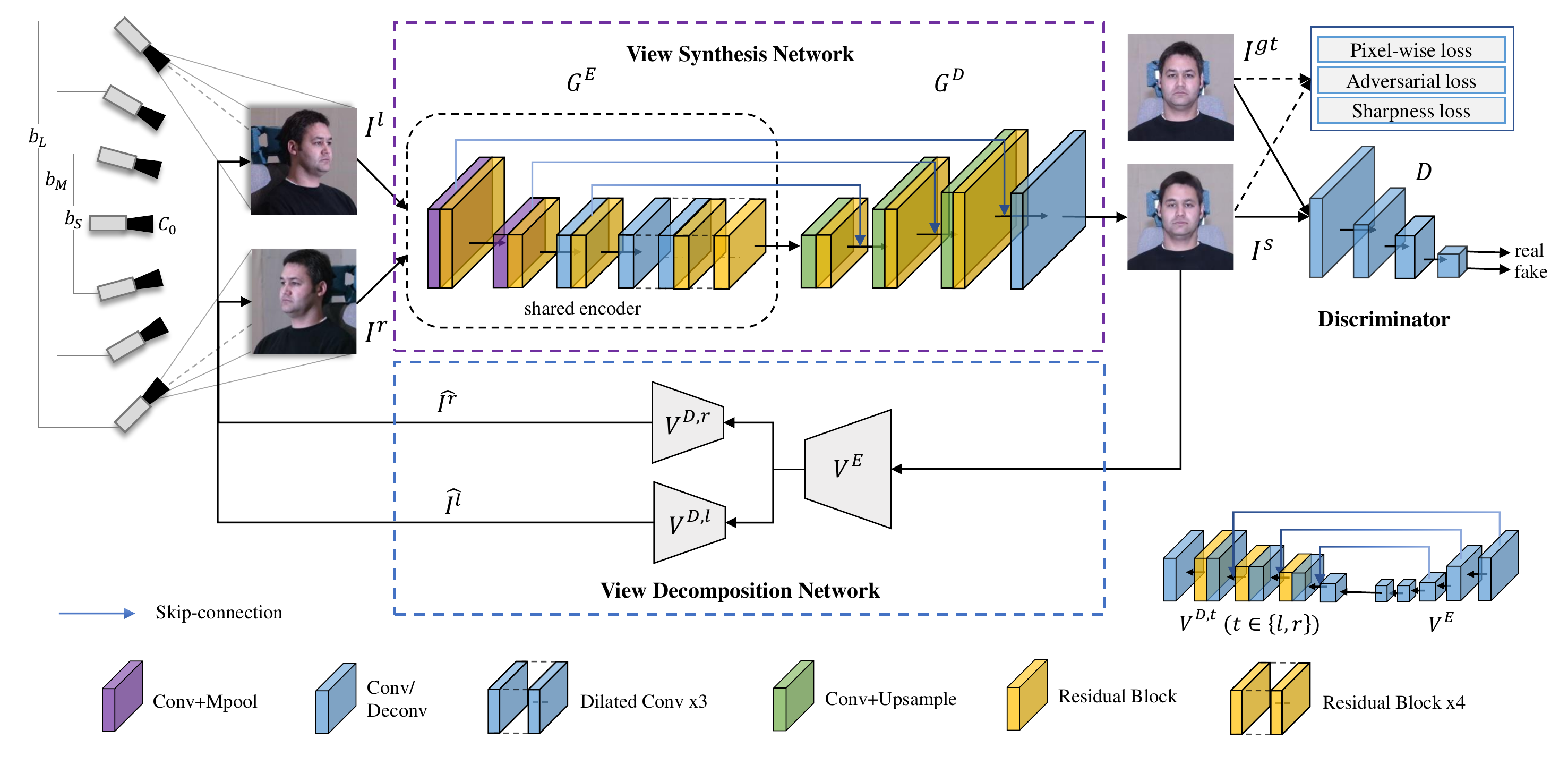}
			\caption{
				General structure of Self-Consistent Generative Network (\SexyName) for the task of \textbf{frontal view synthesis}.  
				Here, camera $C_0$ is the reference camera that produces the ground-truth view image $I^{gt}$. 
				Moreover, the connected pairs of cameras capture symmetrical views with different baselines, where $b_S$, $b_M$ and $b_L$ are view pairs with small, medium, and large baselines, respectively. 
				\SexyName\ seeks to recover the middle view from a given side view pair $(I^l,I^r)$. 
				\SexyName\ consists of two major components, namely the encoder-decoder-based generator $\{G^E, G^D\}$, and the view decomposition network which enforces consistency between input views and recovered views. 
				More network details can be found in Section~\ref{sec:network}.
			}
			\label{fig:sagan}
		\end{center}
		\vspace{-0.1in}
	\end{figure*}	
	
	Our main contributions are summarized as follows.
	\begin{itemize}
		\item
		We propose a novel deep View Synthesis Network, called Self-consistent Generative Network (\SexyName), which simultaneously synthesizes photo-realistic unseen views and preserves high consistency among different views of the same scene.
		\item 
		We propose a View Decomposition Network (VDN) that reconstructs the input views from the synthesized view. 
		In this way, different views are highly correlated with each other and the geometric pre-processing (\eg, rectification) in existing methods becomes not necessary.
		\item
		Comprehensive experiments demonstrate the superior performance of the proposed method over existing methods both quantitatively and qualitatively. 
		In particular, the proposed method is able to produce visually promising middle views on both the benchmark datasets and the real-world conferencing system\footnote{The collected dataset of the real-world conferencing is available at \href{https://zhuomanliu.github.io/datasets/download.html}{https://zhuomanliu.github.io/datasets/download.html}.}.
	\end{itemize}

	
	\section{Related Work}
	\label{sec:relatedWorks}
	{\noindent\textbf{Multi-view synthesis.} Synthesizing a novel view from multiple view images has long been studied. 
		Debevec \etal \cite{debevec1996modeling} combines both image-based and geometry-based techniques to render novel views from multiple views.
		Sagonas \etal \cite{sagonas2015robust} considers frontal facial image synthesizing as an optimization problem. 
		Traditional methods fail in occlusion situations and may generate artifacts in synthesized views. 
		Thus, some approaches that combine different learning methods are proposed to tackle such bottlenecks.
		
		Learning-based approaches tackle multi-view synthesis task via training a prediction model, \eg, Convolutional Neural Networks (CNNs)~\cite{dosovitskiy2015learning,zhou2016view,liu2018geometry,Flynn2016}.
		Dosovitskiy \etal \cite{dosovitskiy2015learning} trains CNN to render images of chairs with different poses, lighting, \etc.
		DeepStereo~\cite{Flynn2016} synthesizes a novel view by interpolating from neighboring posed views of a scene. However, it is difficult to composite occluding content under large baselines. 
		Similarly, Multi2Novel~\cite{sun2018multi} and StereoMagnification~\cite{zhou2018stereo} use multi-plane or multi-view to construct a plane-sweep volume. They have the same problem as DeepStereo.
		Considering the correlation among different views, Zhou \etal \cite{zhou2016view} proposes View Synthesis by Appearance Flow (VSAF) to synthesize new images of the same object from arbitrary viewpoints.
		However, VSAF requires viewpoint transformation information (in addition to the input images) and may lead to incorrect content due to occlusion.
		Park \etal \cite{park2017transformation} and Ji \etal \cite{Ding2017} seek to improve VSAF by addressing these problems.
		Disocclusion-aware Appearance Flow Network (DOAFN)~\cite{park2017transformation} is proposed to predict not only a novel view but also a visibility map to improve performance.
		{
			Multi-Scale Adversarial Correlation Matching (MS-ACM)~\cite{Zhang_2019_CVPR} models structures as self-correlation coefficients extracted from multi-scale feature maps.
		}
		Unfortunately, both DOAFN and MS-ACM are not suitable for the multi-view synthesis task studied in this paper due to the limitation of a single view input.
		{
			To render a novel view,  View Independent Generative Adversarial Network (VI-GAN)~\cite{Xu_2019_ICCV} and Extreme View Synthesis \cite{Choi_2019_ICCV} input additional camera pose, which is not required in our multi-view synthesis task.
		}
		DVM \cite{Ding2017} aims to synthesize novel views from multiple views without additional information beyond input image pairs. 
		Our model outperforms DVM when dealing with a large baseline image pair.
		
		\noindent\textbf{Generative adversarial networks (GANs).} Recently, many GANs~\cite{goodfellow2014generative,guo2019auto,cao2020improving} have been proposed to generate images, such as DCGANs \cite{radford2015unsupervised}, WGANs \cite{arjovsky2017wasserstein}, and progressive GANs \cite{karras2017progressive}.
		Inspired by GANs, Huang \etal \cite{huang2017} proposed a Two-Pathway Generative Adversarial Network (TP-GAN) to synthesize a facial view image from one side view while preserving the symmetric structure of faces.
		However, TP-GAN ignores data consistency and may result in meaningless images. 
		Regarding this issue, Zhu \etal proposed cycle-consistency loss to preserve the content in image translation by enforcing double-sided consistency during training~\cite{zhu2017unpaired,guo2020closed}.
		Better than Pix2Pix~\cite{isola2017image}, the double forward-backward processes qualify it for unsupervised tasks. 
		However, unlike image translation, the multi-view synthesis task is often a supervised task in which we should exactly recover a novel view from two or more view images (or videos) obtained by cameras at distinct positions with more strict constraints.
		Focusing on addressing the multi-view synthesis task, in this paper, we use self-consistency (one cycle mapping from the synthesized view to side views) to ensure the input views (which may contain occluded contents) can be reconstructed.

		
		\section{Self-Consistent Generative Network}
		\label{sec:method}
		With the goal of addressing the challenges in the view synthesis task, \ie, synthesizing novel views under large baselines with more occluded areas in paired views, and avoiding the limitations of geometric modules under large baselines, we propose self-consistency in our model.
		
		Without loss of generality, we introduce our method by focusing on synthesizing the absent frontal view from two stereo views, as shown in Fig. \ref{fig:comp_intro}.
		Given a set of stereo view triplets $\{(I_i^l,I_i^r, I_i^{gt})\}_{i=1}^{n}$, we seek to learn a mapping $G: (I^{l},I^{r})\to I^{gt}$ to recover the ground-truth view $I^{gt}$ from a given input view pair $(I^l,I^r)$. 
		This task is non-trivial due to the view correspondence issue of distinct views.
		
		In this paper, we present a novel view synthesis method, called Self-Consistent Generative Network (\SexyName).
		As shown in Fig.~\ref{fig:sagan}, our proposed method consists of two parts, namely a \emph{View Synthesis Network} (VSN) for generating a frontal view from two stereo views, and a \emph{View Decomposition Network (VDN)}  for attempting to reconstruct two input stereo views from the synthesized frontal view.
		Here, VDN helps to address the occlusion problem caused by a large baseline. 
		Furthermore, to ensure the photo-realism of the synthesized views, we further introduce a GAN based loss to train the model rather than the simple pixel-wise loss. 
		The details of each part will be described in the following sections.
		
		\subsection{View Synthesis Network}
		\label{sec:VSN}
		As shown in the purple dotted block of Fig.~\ref{fig:sagan},	we employ an encoder-decoder network to implement the view synthesis network $G$, consisting of an encoder $G^E$ and a decoder $G^D$.
		Both the encoder and the decoder networks are composed of a stack of residual blocks~\cite{he2016deep}, allowing for faster convergence and better performance. 
		To improve the representation ability of the embedding, in the encoder, we replace the last three convolutional layers with dilated convolution~\cite{yu2015multi, li2019dilated} to increase the receptive field of the filters without increasing the number of weights.
		
		Given the left and right views $I^l$ and $I^r$, the synthesized view, denoted by $I^s$, can be computed by
		\begin{equation}\label{eq:synthesize_frontal_views}
		I^s = G^D \left( G^E (I^l, I^r) \right),
		\end{equation}
		where $G^E(I^l, I^r)$ denotes feature extraction from $I^l$ and $I^r$ using a weight-shared encoder. Specifically, we first use the encoder model to extract features from $I^l$ and $I^r$ independently. Then, we concatenate the features of $I^l$ and $I^r$ as the output of $G^E(I^l, I^r)$.

		\noindent\textbf{Reconstruction loss.} 
		To exactly recover the frontal view image, it is straightforward to use a pixel-wise loss to minimize the distance between the synthesized view $I^s$ and the ground-truth $I^{gt}$ on pixel level:}
	\begin{equation}
	L_p(\theta_G)= \frac{1}{n} \sum_{i=1}^{n} \|I^s_{i}-I^{gt}_{i}\|_1,
	\end{equation}
	where $n$ denotes the number of images, $\|\cdot\|_1$ denotes $\ell_1$-norm.
	
	\noindent\textbf{Sharpness loss.} To improve the quality of the synthesized images, we integrate an image sharpness method into the loss function. First, to measure the sharpness of images, we exploit the sharpness criterion $Q_S$ in LOGS~\cite{quality2018} by computing the differences of the textural complexity between the synthesized image and its reblurred version obtained by a Gaussian smoothing filter.
    The textural complexity can be represented by the standard deviations of the pixels in the image. Following~\cite{quality2018}, we compute $Q_S$ in a block-wise manner:
        \begin{equation}\label{eq:logs_Qs}
            Q_S(I) = \frac{\sum_{i=1}^{i=Z}\sqrt{|\sigma_{1i}^2-\sigma_{2i}^2|}}{Z},
        \end{equation}
    where $\sigma_{1i}^2$ and $\sigma_{2i}^2$ represent the standard deviations of the $i$-th block in the image and its blurred version. Here, $Z=\lfloor \frac{M}{k}\rfloor\cdot\lfloor \frac{N}{k}\rfloor$ denotes the total number of blocks, where $k$ denotes the block size, $M$ and $N$ denote the height and width of the image $I$, respectively. 
    Then, we construct the sharpness loss $L_{sharp}$ based on $Q_S$:
    \begin{equation}\label{eq:sharpness}
        L_{sharp} = \frac{1}{n}\sum_{i=1}^{n} \|Q_S(I_i^{gt}) - Q_S(I_i^s)\|_1,
    \end{equation}
    where $I^s$ denotes the synthesized view and $I^{gt}$ denotes the ground-truth view.

	\noindent\textbf{Adversarial loss.}
	To improve the photo-realism of the synthesized views, we propose to train the network in an adversarial manner. 
	Specifically, VSN can be regarded as a generator $G$ for synthesizing a frontal view $I^s$ that is as photo-realistic as the real one $I^{gt}$. 
	To enable VSN to synthesize a good-quality frontal view, we also introduce discriminator $D$ to distinguish the generated frontal view from a real frontal view.

	\begin{algorithm}[t]
		\caption{Training algorithm for \SexyName}
		\begin{algorithmic}[1]
			\Require Training stereo view triplets $\{I^l_i, I^r_i, I^{gt}_i\}_{i=1}^n$; batch size $m$; number of training iterations $T$; learning rate $\alpha$.
			\For{$t=1,...,T$} \label{alg:train}
			\State Sample a mini-batch of views $\{I^l_i, I^r_i, I^{gt}_i\}_{i=1}^{m}$.
			\State Synthesize the frontal view $I_i^s$ using Eq.~(\ref{eq:synthesize_frontal_views}).
			\State Synthesize view pair $(\hat{I^l_i},\hat{I^r_i})$ from $I_i^s$ using Eq.~(\ref{eq:decom}).
			\State Update discriminator parameters $\theta_D$ using Eq.~(\ref{eq:update_theta_D}).
			\State Update VSN parameters $\theta_G$ using Eq.~(\ref{eq:update_theta_G}).
			\State Update VDN parameters $\theta_V$ using Eq.~(\ref{eq:update_V}).
			\EndFor
		\end{algorithmic}
	\end{algorithm}

	Let $\theta_G$ and $\theta_D$ be the model parameters of the generator $G$ and the discriminator $D$, respectively.
	Following~\cite{goodfellow2014generative}, the adversarial network can be trained by solving the following minimax problem:
	\begin{equation}\label{eq:optimization_problem}
	 \min_{\theta_G} \max_{\theta_D} L_{gen}(\theta_G, \theta_D),
	\end{equation}
	with $L_{gen}(\theta_G, \theta_D)$ being
	\begin{equation}
	\begin{aligned}\label{eq:3}
	L_{gen}(\theta_G, \theta_D) &= \mathbb{E}_{I^{gt}\sim P_{I^{gt}}}[\log D(I^{gt})]\\
	&+\mathbb{E}_{I^s\sim P_{I^s}}[\log(1-D(I^s))],
	\end{aligned}
	\end{equation}
	where $P_{I^{gt}}$ and $P_{I^s}$ are the distributions of the ground-truth and synthesized image, respectively. 
	
	In the training,
	the discriminator $D$ can be learned by minimizing the following loss:
	\begin{equation}\label{eq:2}
	L_{disc}(\theta_D) = -\frac{1}{n} \sum_{i=1}^n \log D(I_i^{gt})-\log(1-D(I_i^s)),
	\end{equation}
	where $D \left( I^s \right)$ is the probability that a synthesized image is a real frontal view. 
	For better gradient behavior, we minimize $-\log D \left( I^s \right)$ instead of $\log \left( 1 - D \left( I^s \right) \right)$~\cite{goodfellow2014generative}.
	For the generator $G$, we can define the \textbf{adversarial loss} as follows:
	\begin{equation}\label{eq:loss_adv}
	L_{adv}(\theta_G) = - \frac{1}{n}\sum_{i=1}^n \log D \left(G^D \left(G^E (I_i^l, I_i^r) \right) \right).
	\end{equation}
	During training, the adversarial loss will be combined with other losses to update the parameters $\theta_G$ of generator $G$.

	\subsection{View Decomposition Network}
	\label{sec:VDN}
	For the ill-posed problem that there exists an infinite number of middle/novel views that correspond to the same input views~\cite{huang2017}, we propose a self-consistency scheme to reduce the space of possible view synthesis functions. Specifically, we propose a View Decomposition Network (VDN) that reconstructs the input side views from the predicted middle/novel view, as shown in the blue block of Fig. \ref{fig:sagan}.
 
	The VDN consists of an encoder $V^E$ and two separate decoders $V^D$ for the two side views.
	Specifically, VDN introduces a decomposition mechanism to decompose the generated frontal image $I^s$ from VSN backward into $(\hat{I^l},\hat{I^r})$, \ie,
	\begin{equation}\label{eq:decom}
	\hat{I^l}=V^{D,l}(V^E(I^s)) \quad \mathrm{and~~}
	\hat{I^r}=V^{D,r}(V^E(I^s)).
	\end{equation}
	
	In addition, by regenerating the side views, VDN here can ensure the validity of the generated occluded area. 
	This is the reason why the VSN in our model does not need to contain a rectification module or any transformation operations.
	In combination with the forward generation network (\ie, VSN) that learns the translation from $(I^l,I^r)$ to $I^s$, VDN can backtrack to the original source and enforce forward-backward constraints on input view pairs. 
	The predicted left and right views $(\hat{I^l},\hat{I^r})$ should be close to the real left and right input $(I^l,I^r)$. 
	We, therefore, minimize the distance between $(\hat{I^l},\hat{I^r})$ and $(I^l,I^r)$ through $L_{vc}$:
	\begin{equation}\label{eq:synthesize_left_right_views}
	L_{vc}(\theta_G,\theta_V) =\frac{1}{n}\sum_{i=1}^{n}\ \|\hat{I_i^l}-I_i^l\|_1+\|\hat{I_i^r}-I_i^r\|_1.
	\end{equation}
	
	Noted that the proposed self-consistency has several difference with cycle-consistency~\cite{zhu2017unpaired}. Firstly, cycle-consistency uses cycles to help minimize distribution divergence without ground truth while our self-consistency builds a cycle to improve the pixel-wise prediction together with reconstruction loss. Secondly, cycle-consistency learns two symmetric mappings between the images in two domains while our self-consistency learns two asymmetric mappings, \emph{i.e.}, a synthesis mapping and a decomposition mapping. In practice, the proposed self-consistency scheme is able to significantly improve the performance by incorporating the constraint \wrt the decomposition mapping (See Table~\ref{tab:error_small}, Table~\ref{tab:error_large}, and Table~\ref{tab:error_kitti} in the paper).

	\subsection{Training Details}
	\label{sec:trainDetails}
	To train \SexyName, we need to update the parameters $\theta_G$ for VSN, $\theta_V$ for VDN and $\theta_D$ for discriminator $D$. 
	Following GANs~\cite{goodfellow2014generative}, we adopt an alternating optimization scheme to train \SexyName\ using mini-batch stochastic gradient descent (SGD), as shown in Algorithm~\ref{alg:train}. 
	Let $\alpha$ be the learning rate for SGD. 
	In each iteration, for discriminator $D$, we update $\theta_D$ by minimizing the loss $L_{disc}$ according to
	\begin{equation}
	\label{eq:update_theta_D}
	\theta_D = \theta_D - \alpha \nabla_{\theta_D} L_{disc}.
	\end{equation} 
	For VSN, we should update $\theta_G$ by minimizing the following loss function:
	\begin{equation}
	\label{eq:loss_g}
	L_G(\theta_G){=}L_p + \lambda_1 L_{vc} + \lambda_2 L_{adv} + \lambda_3 L_{sharp},
	\end{equation}
	where $\lambda_1$, $\lambda_2$, and $\lambda_3$ are balancing parameters. Specifically, we consider the loss function with the $L_{sharp}$ term as a variant of the loss function without the $L_{sharp}$. We further discuss this variant in Section~\ref{sec:sharpness}. Thus, the update can be made by 
	\begin{eqnarray}
	\label{eq:update_theta_G}
	\theta_G = \theta_G - \alpha \nabla_{\theta_G} L_G.
	\end{eqnarray}
	Last, for VDN, we update $\theta_V$ by minimizing the loss $L_{vc}$  according to
	\begin{equation}
	\label{eq:update_V}
	\theta_V = \theta_V - \alpha \nabla_{\theta_V} L_{vc}.\\
	\end{equation}
	
	\subsection{Details of the Network Structure}
	\label{sec:network}
	
	\subsubsection{View synthesis network}
	\label{sec:vsn}
	We build the view synthesis network $G$ following the scheme of an encoder-decoder network.
	The details are shown in Table \ref{tab:mvg}.
	
	\noindent\textbf{Encoder.}
	The weight-shared encoder $G^E$ takes the left and right views as inputs respectively. For the encoder $G^E$, each convolutional layer is followed by a leaky rectified linear unit (leaky ReLU). To better leverage spatial information and large distance information, dilated convolution is also employed in the encoder. Then, the encoded features of the left and right views $\{ec6\_l, ec6\_r\}$ are concatenated along the channel dimension and taken as the inputs of the decoder $G^D$. We also introduce residual blocks to our model. Specifically, each max-pooling layer is followed by one residual block and the final layer of the encoder (\ie, ec6) is followed by four residual blocks.
	
	\noindent\textbf{Decoder.}
	For the decoder $G^D$, we adopt ReLU as the non-linear activation function after each convolutional layer except for the final convolution layer $dc5$. In layer $dc5$, tanh is adopted to keep the output within the normalized data range.
	We obtain $\{ecfeatk\_l, ecfeatk\_r\}$ ($k\in\{1,2,3\}$) by applying $1\times 1$ kernels to the outputs of $\{eck\_l, eck\_r\}$ ($k\in\{1,2,3\}$), respectively. We insert skip-connection between the encoder and the decoder and obtain the input of the next layer by concatenating $\{ecfeatk\_l, ecfeatk\_r\}$ ($k\in\{1,2,3\}$) with $upk$ ($k\in\{1,2,3\}$).

	\subsubsection{View decomposition network}
	\label{sec:vdn}
	Unlike the view synthesis network $G$, the view decomposition network $V$ is a fully convolutional network with the structure shown in Table \ref{tab:vcs}.
	
	It is noteworthy that, we obtain $\{dec5\_l, dec5\_r\}$ by applying two $1\times 1$ kernels to the output of $dec5$ in the encoder $V^E$. Then, the two decoders individually process $\{dec5\_l, dec5\_r\}$ to acquire their decomposed views. Leaky ReLU, residual blocks, and skip-connections are introduced in this network, similar to the view synthesis network, to ensure the effectiveness of our model.

	\subsubsection{Discriminator network}
	\label{sec:disc}
	We show the detailed structure of the discriminator network $D$ in Table \ref{tab:fvd}. Each convolutional layer is followed by a leaky ReLU. 
	The fully connected layer on top of the convolutional layers is used to estimate the probability that the middle view $I^s$ or $I^{gt}$ is real.
	
	\begin{table}[tp]
		\caption{Detailed structure of the view synthesis network. The layer types ``conv'', ``maxpool'', ``dconv'', and ``upsample'' represent ``convolution'', ``max-pooling'', ``dilated convolution'' and ``upsampling'' respectively. $k$ denotes the kernel size, $s$ is the stride of the layer and $r$ denotes the dilation rate of dilated convolution. The default input of each layer is the output of the previous layer, except for those layers specified by the column ``Input''.}
		\label{tab:mvg}
		\small
		\begin{center}
			\begin{minipage}[b]{1\linewidth}
				\centering
				\resizebox{0.7\textwidth}{!}{
					\begin{tabular}{c|c|c|c|c|c}
						\hline
						\multicolumn{6}{c}{\textbf{Shared Encoder}}\\
						\hline
						Layer & Type & $k$ & $s$ & $r$ & Output Size \\
						\hline
						ec1 & conv & 7 & 1 & - & 224$\times$224$\times$32 \\
						ep1 & maxpool & 3 & 2 & - & 112$\times$112$\times$32 \\
						ec2 & conv & 5 & 1 & - & 112$\times$112$\times$64 \\
						ep2 & maxpool & 3 & 2 & - & 56$\times$56$\times$64 \\
						ec3 & conv & 3 & 1 & - & 56$\times$56$\times$128 \\
						ec4 & dconv & 3 & 1 & 2 & 56$\times$56$\times$128 \\
						ec5 & dconv & 3 & 1 & 2 & 56$\times$56$\times$128 \\
						ec6 & dconv & 3 & 1 & 2 & 56$\times$56$\times$128 \\
						\hline
					\end{tabular}
				}
			\end{minipage}
			\\	\vspace{0.3cm}
			\begin{minipage}[b]{1\linewidth}
				\centering
				\resizebox{1\textwidth}{!}{
					\begin{tabular}{c|c|c|c|c|c}
						\hline
						\multicolumn{6}{c}{\textbf{Decoder}}\\
						\hline
						Layer & Type & Input&$k$ & $s$ & Output Size\\
						\hline
						dc1 & conv & ec6\_l, ec6\_r&3 & 1 & 56$\times$56$\times$128\\
						up1 & upsample &dc1 &- & - & 56$\times$56$\times$128\\
						dc2 & conv & up1, ecfeat3\_l, ecfeat3\_r&3 & 1 & 56$\times$56$\times$64\\
						up2 & upsample & dc2&- & - & 112$\times$112$\times$64\\
						dc3 & conv & up2, ecfeat2\_l, ecfeat2\_r&5 & 1 & 112$\times$112$\times$32\\
						up3 & upsample &dc3 &- & - & 224$\times$224$\times$32\\
						dc4 & conv &up3, ecfeat1\_l, ecfeat1\_r &7 & 1 & 224$\times$224$\times$32\\
						dc5 & conv &dc4 &3 & 1 & 224$\times$224$\times$3\\
						\hline
					\end{tabular}
				}
			\end{minipage}
		\end{center}
	\end{table}
	
	\begin{table}[tp]
		\caption{Detailed structure of the view decomposition network. ``deconv'' represents the deconvolution layer.}
		\label{tab:vcs}
		\small
		\begin{center}
			\begin{minipage}[b]{1\linewidth}
				\centering
				\resizebox{1.0\textwidth}{!}{
					\begin{tabular}{c|c|c|c|c|c|c|c|c|c}
						\hline
						\multicolumn{5}{c|}{\textbf{Encoder}}&
						\multicolumn{5}{c}{\textbf{Decoder}} \\
						\hline
						Layer & Type & $k$ & $s$ & Output Size &
						Layer & Type & $k$ & $s$ & Output Size\\
						\hline
						dec1 & conv & 7 & 2 & 112$\times$112$\times$16 &
						ddc1 & deconv & 3 & 2 & 28$\times$28$\times$128\\
						dec2 & conv & 5 & 2 & 56$\times$56$\times$32 &
						ddc2 & deconv & 3 & 2 & 56$\times$56$\times$64\\	
						dec3 & conv & 3 & 2 & 28$\times$28$\times$64 &
						ddc3 & deconv & 5 & 2 & 112$\times$112$\times$32\\
						dec4 & conv & 3 & 2 & 14$\times$14$\times$128 &
						ddc4 & deconv & 7 & 2 & 224$\times$224$\times$16\\
						dec5 & conv & 3 & 1 & 14$\times$14$\times$256 &
						ddc5 & deconv & 3 & 1 & 224$\times$224$\times$3\\
						\hline
					\end{tabular}
				}
			\end{minipage}
		\end{center}
		\vspace{-0.5cm}
	\end{table}	
	
	\begin{table}[!tp]
		\begin{center}
			\caption{Detailed structure of the discriminator network. }
			\label{tab:fvd}
			\vspace{-0.2cm}
			\resizebox{0.3\textwidth}{!}{
				\begin{tabular}{c|c|c|c|c}
					\hline
					Layer & Type & $k$ & $s$ & Output Size\\
					\hline
					disc1 & conv & 5 & 2 & 112$\times$112$\times$32\\
					disc2 & conv & 5 & 2 & 56$\times$56$\times$64\\	
					disc3 & conv & 5 & 2 & 28$\times$28$\times$128\\
					disc4 & conv & 5 & 2 & 14$\times$14$\times$256\\
					fc5 & fc & - & - & 1\\
					\hline
				\end{tabular}
			}
		\end{center}
		\vspace{-0.5cm}
	\end{table}

	
	\section{Experiments}
	\label{sec:exp}
	To demonstrate the effectiveness and robustness of the proposed method, we compare \SexyName\ with several state-of-the-art methods in both indoor and outdoor scene synthesis settings.
	Specifically, we conduct multi-view synthesis experiments on Multi-PIE~\cite{GROSS2010807} and KITTI~\cite{Geiger2012} datasets for the indoor and outdoor scene synthesis tasks, respectively. 
	
	We also apply our method in real conferencing systems\footnote{The demos and the implementation of the proposed \SexyName\ are available at https://github.com/zhuomanliu/\SexyName.}. We collect frames (roughly $5K$ triplets) containing multiple human subjects with their upper bodies in a conferencing scenario. We use $80\%$ for fine-tuning and $20\%$ for testing. The data for fine-tuning and testing share the same backgrounds.
	The same subject with different clothing or motions may appear in different frames. Thus, there is no overlap in image level data.
	More details of the data collection are described in Section~\ref{sec:demo}.

	\begin{figure*}[!htbp]
		\centering
		\subfigure[Training PSNR]{\label{fig:train_psnr}
			\centering
			\includegraphics[trim = 1mm 1mm 8mm 6mm,
			clip, width=0.23\linewidth]{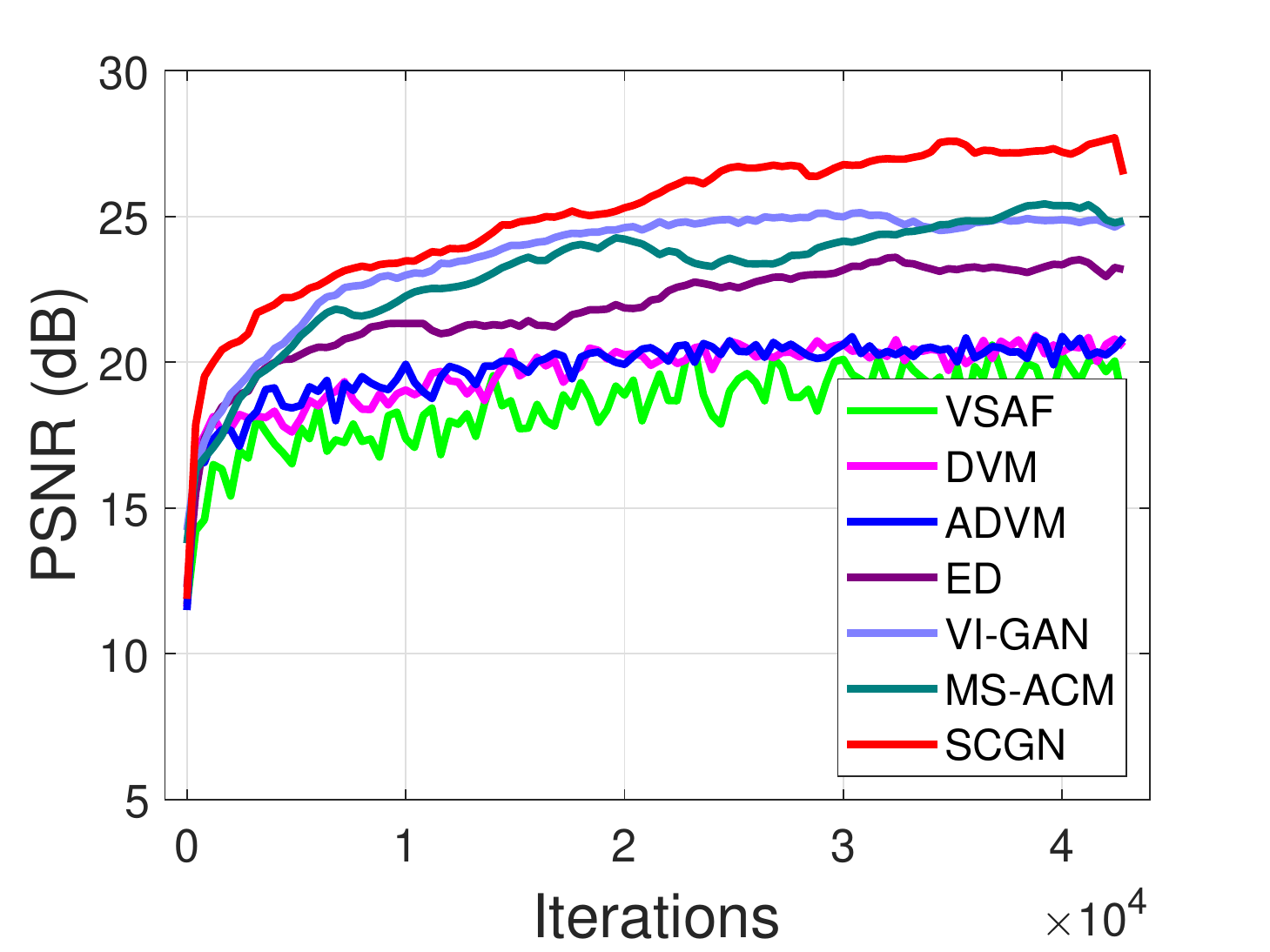}
		}
		\subfigure[Testing PSNR]{\label{fig:test_psnr}
			\centering
			\includegraphics[trim = 1mm 1mm 8mm 6mm,
			clip, width=0.23\linewidth]{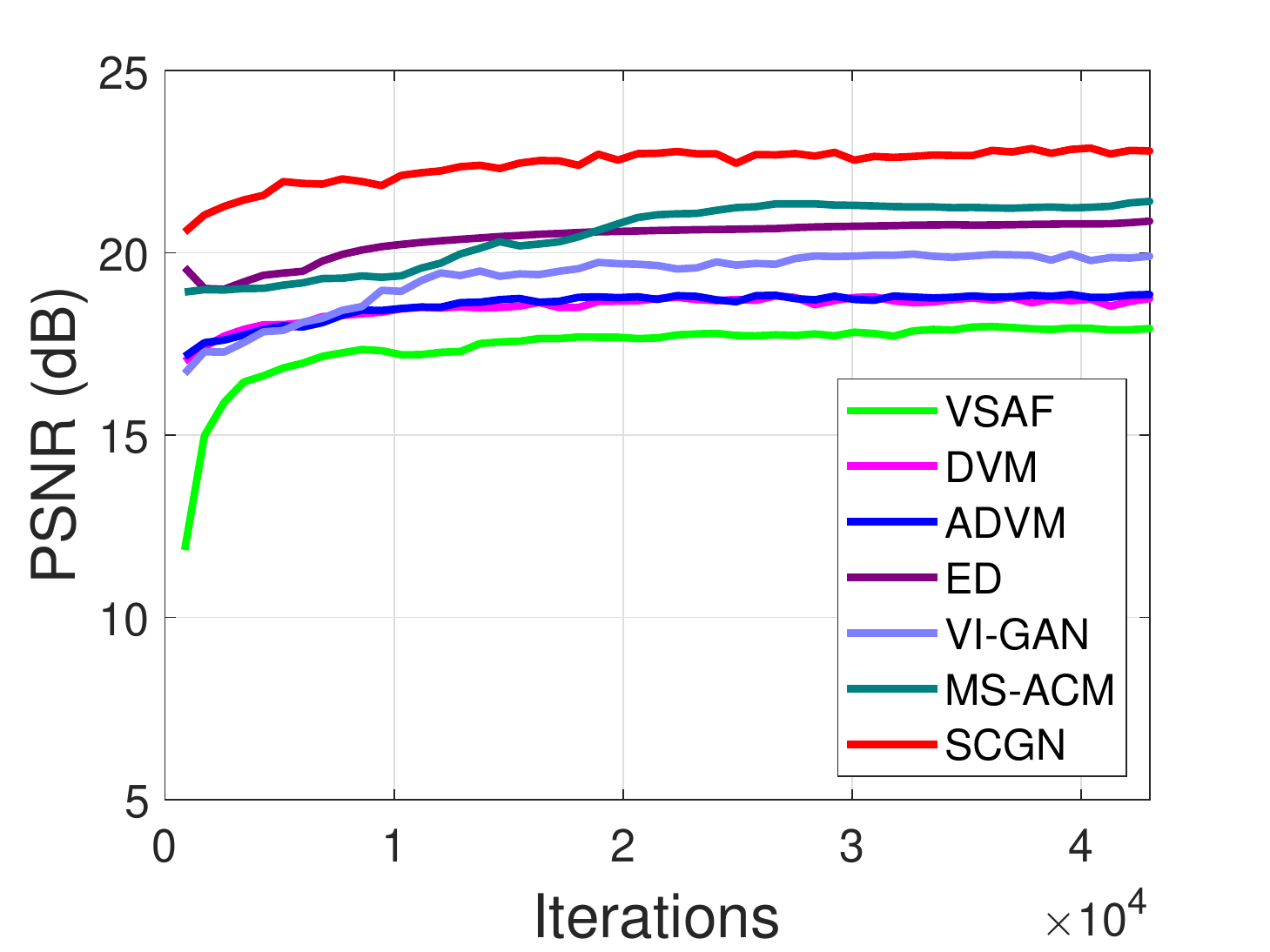}
		}
		\subfigure[Training MS-SSIM]{\label{fig:train_msssim}
			\centering
			\includegraphics[trim = 1mm 1mm 8mm 6mm,
			clip, width=0.23\linewidth]{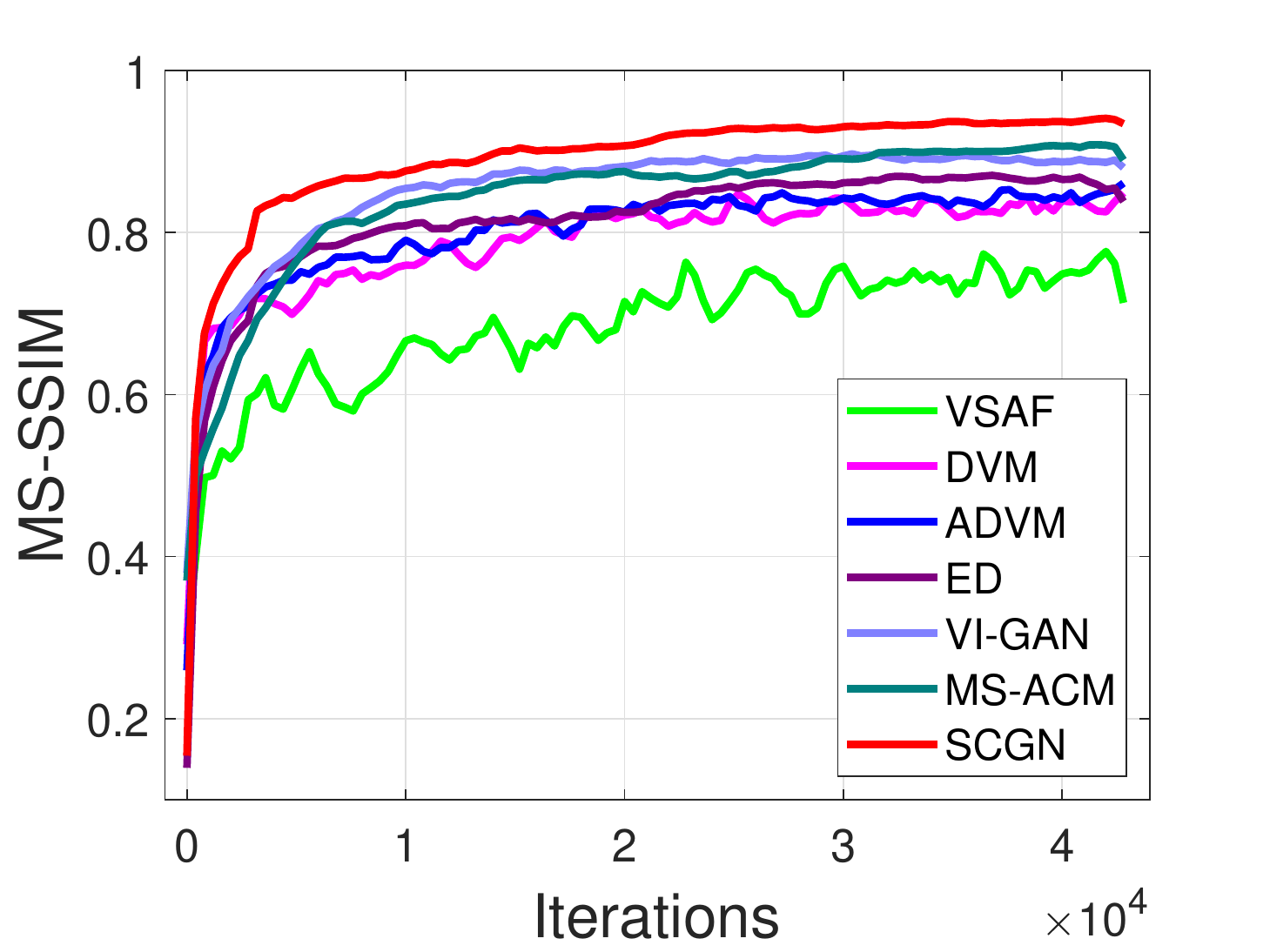}
		}
		\subfigure[Testing MS-SSIM]{\label{fig:test_msssim}
			\centering
			\includegraphics[trim = 1mm 1mm 8mm 6mm,
			clip, width=0.23\linewidth]{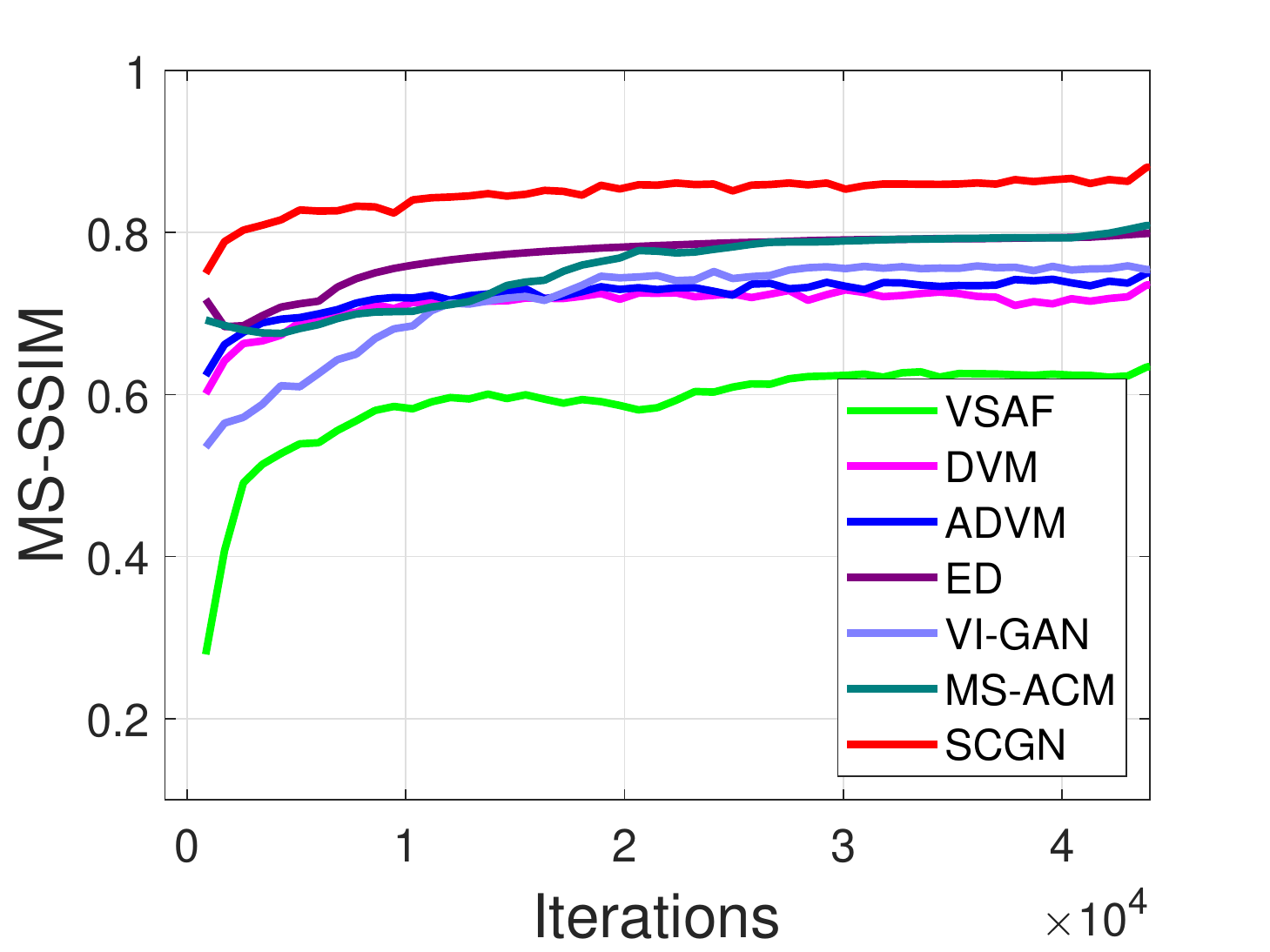}
		}		
		\caption{Performance comparison of different view synthesis methods on Multi-PIE measured by PSNR and MS-SSIM metrics. Both the training and testing curves are reported in this figure.} 
		\label{fig:psnr_msssim}
	\end{figure*}
	
	\begin{figure*}[!htbp]
		\centering
		\subfigure[Training PSNR]{\label{fig:train_psnr_kitti}
			\centering
			\includegraphics[trim = 1mm 1mm 8mm 6mm,
			clip, width=0.23\linewidth]{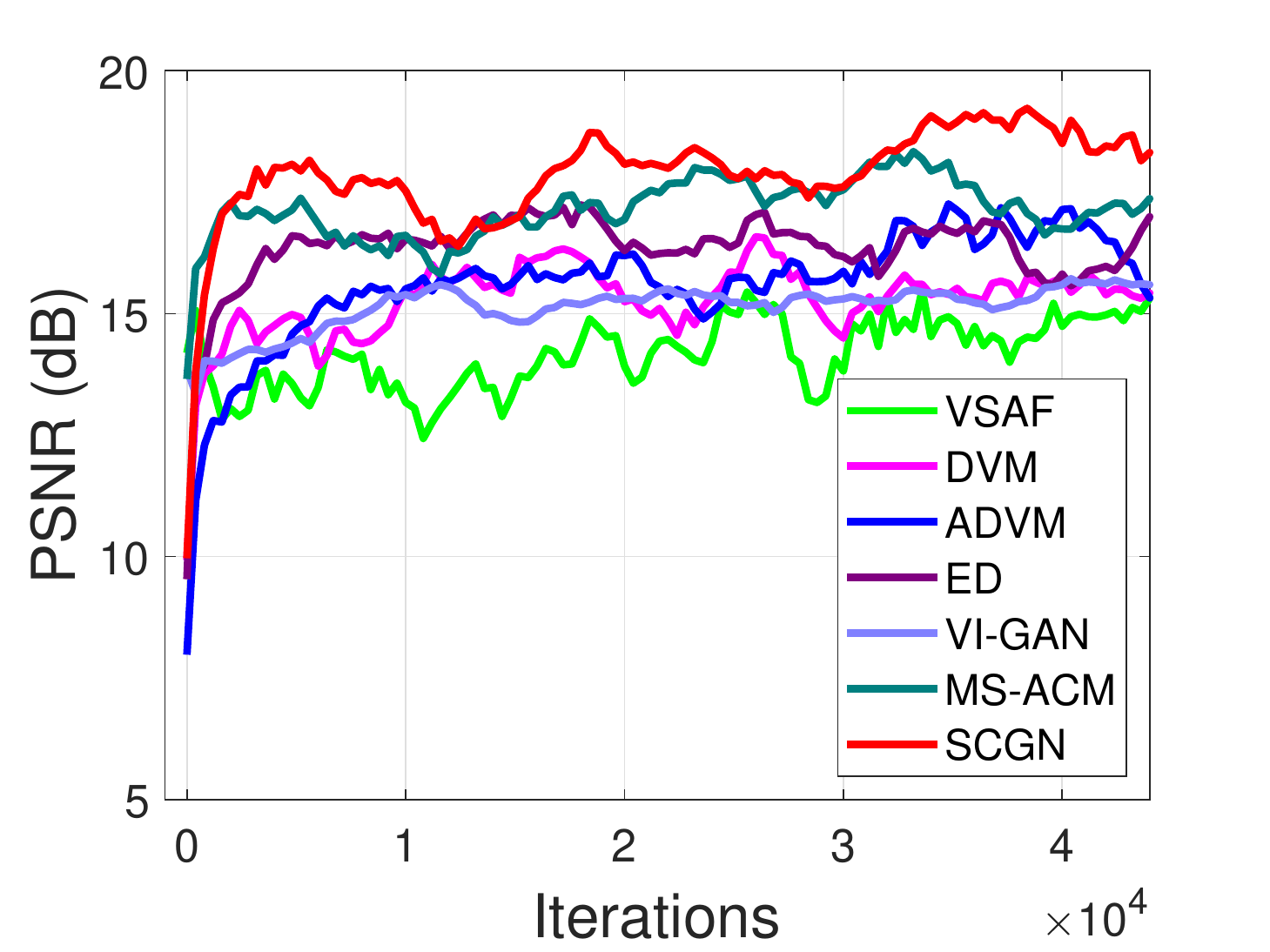}
		}
		\subfigure[Testing PSNR]{\label{fig:test_psnr_kitti}
			\centering
			\includegraphics[trim = 1mm 1mm 8mm 6mm,
			clip, width=0.23\linewidth]{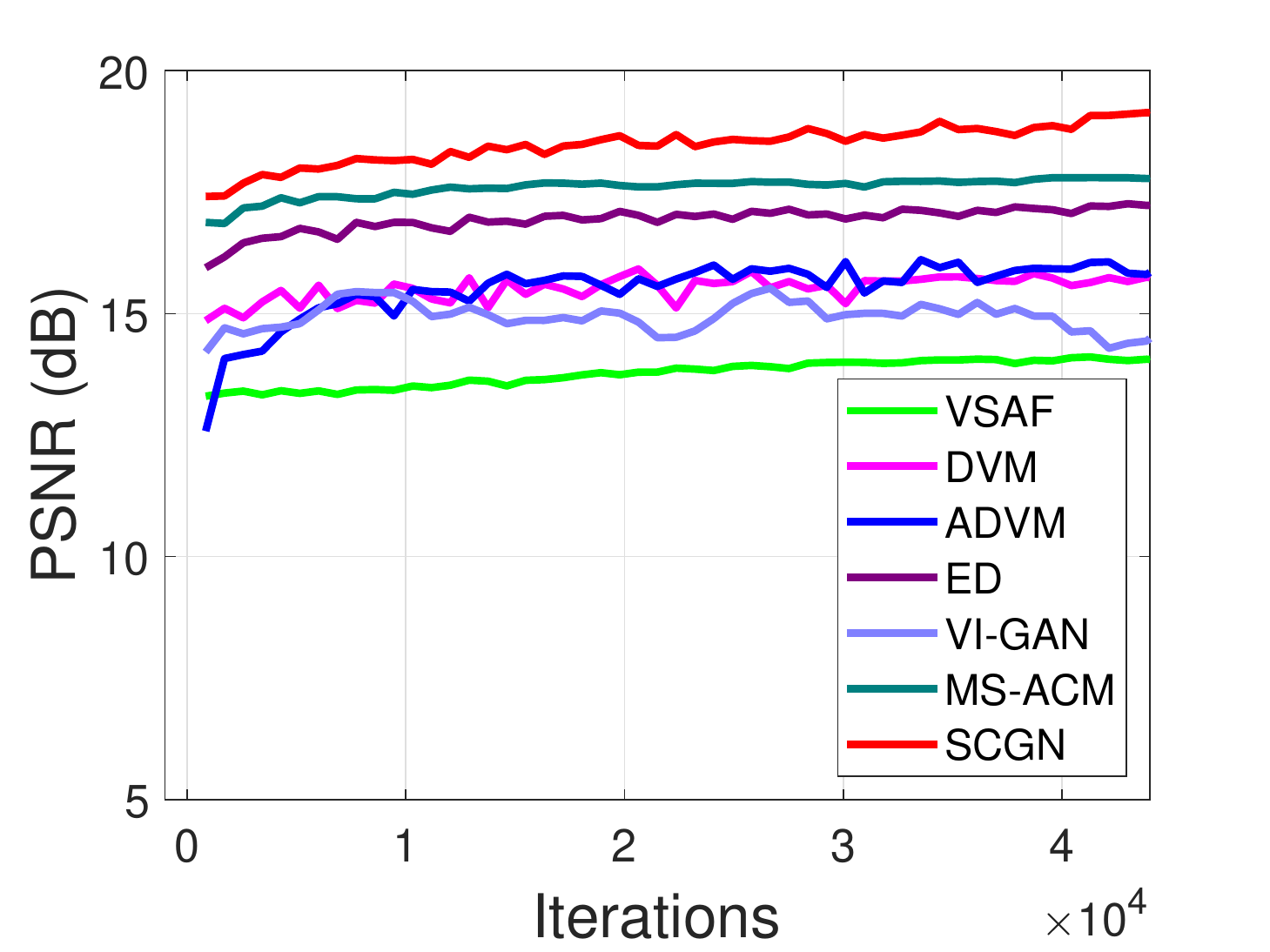} 		
		}
		\subfigure[Training MS-SSIM]{\label{fig:train_msssim_kitti}
			\centering
			\includegraphics[trim = 1mm 1mm 8mm 6mm,
			clip, width=0.23\linewidth]{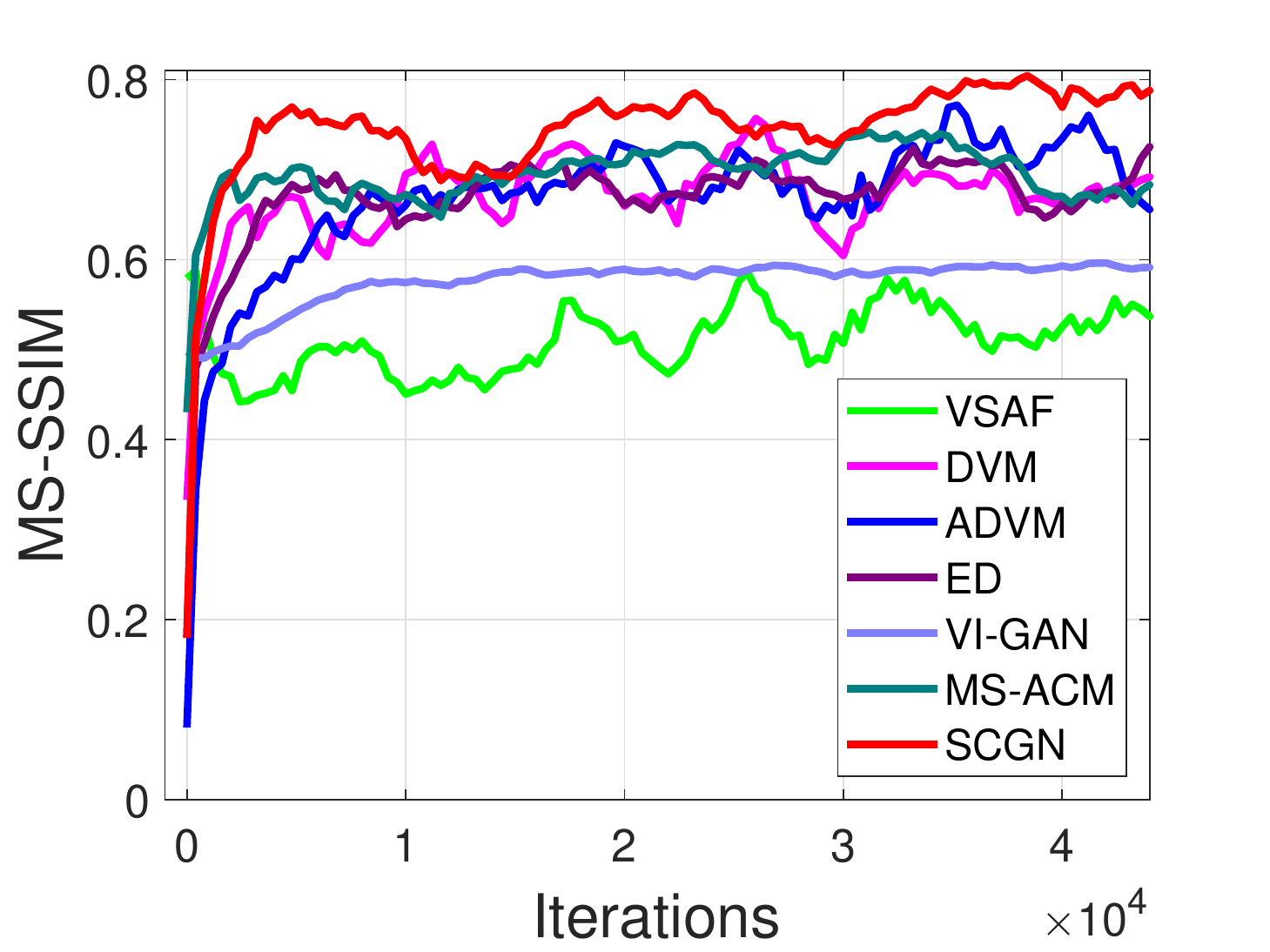}
		}
		\subfigure[Testing MS-SSIM]{\label{fig:test_msssim_kitti}
			\centering
			\includegraphics[trim = 1mm 1mm 8mm 6mm,
			clip, width=0.23\linewidth]{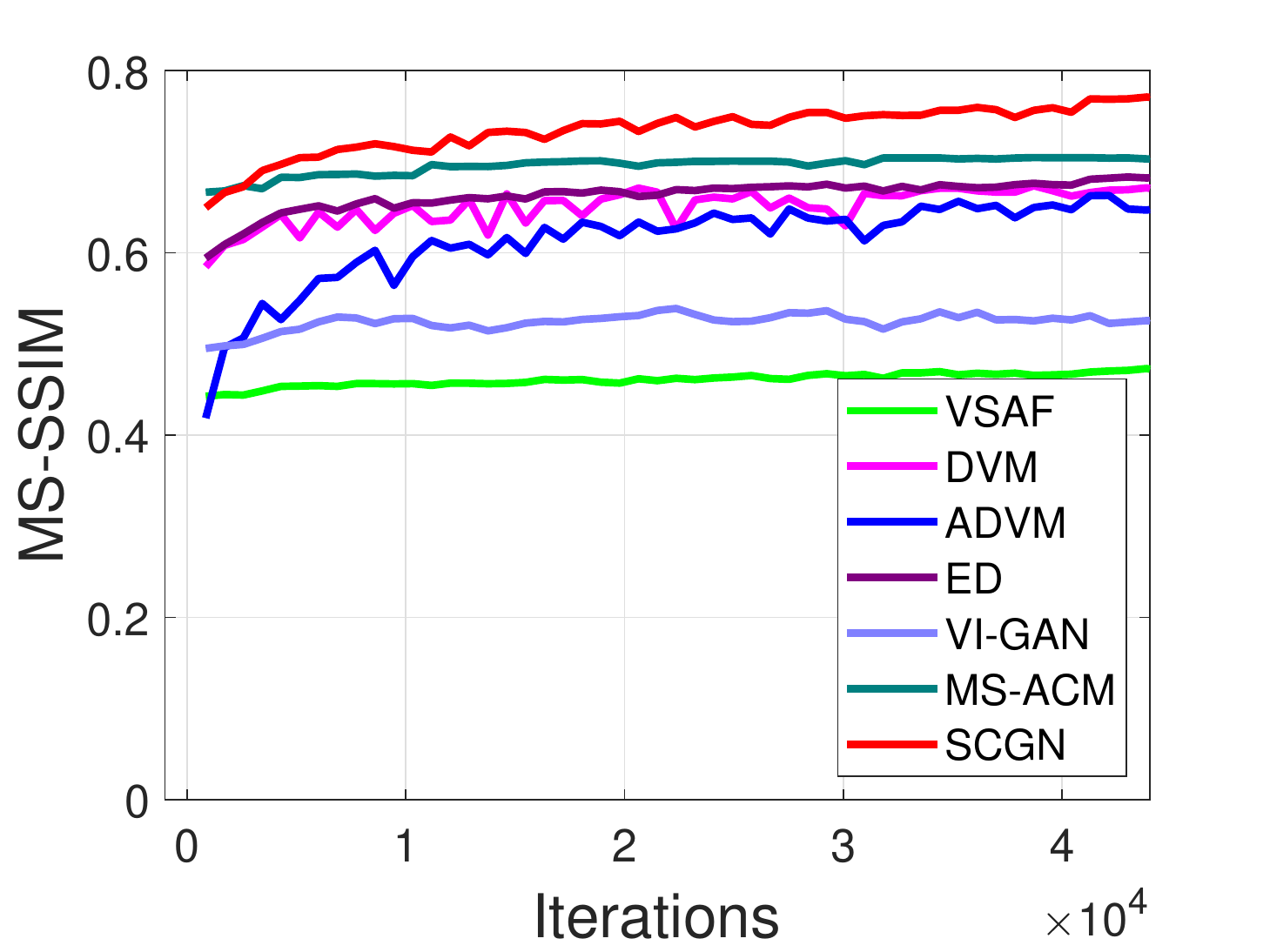}
		}		
		\caption{Performance comparison of different view synthesis methods on KITTI measured by PSNR and MS-SSIM metrics. Both the training and testing curves are reported in this figure.} 
		\label{fig:psnr_msssim_kitti}
	\end{figure*}

	\subsection{Implementation Details and Datasets}
	\label{exp:implementation} 
	For convenience, we use the same experimental settings for experiments on both Multi-PIE and KITTI datasets.
	Specifically, for the optimization, we use Adam~\cite{kingma2014adam} with $\beta_1=0.9$ and $\beta_2=0.999$ and normalize all images to the range $(-1, 1)$
	to train the model 
	while set $\lambda_1=0.01$, $\lambda_2=0.001$ and $\lambda_3=0.01$ to balance the losses in Eq.(\ref{eq:loss_g}). We train the proposed \SexyName~for $371,400$ iterations with a batch size of 1. The learning rate is set to $1\times 10^{-4}$ and $1\times 10^{-5}$ for generator and discriminator, respectively, and decays by $0.1$ at iteration $185,700$. During training, the original images are center cropped to $\textit{min}(H, W)\times\textit{min}(H, W)$, where $H$ and $W$ indicate the height and width of the original image, respectively. Then, we resize it to $224\times 224$.
	We implement our proposed method on TensorFlow~\cite{abadi2016tensorflow} and conduct all experiments on a single Nvidia TitanX GPU.

	\noindent\textbf{Multi-PIE.}
	The indoor dataset Multi-PIE contains more than $750,000$ images of $337$ people under $15$  symmetrical viewpoints.
	For all of the experiments on Multi-PIE in this paper, we divide the dataset into 270 people for training and 67 people for testing, and make image triplets that include the frontal view (\ie, the $0^\circ$ pose captured by the central camera
	\footnote{http://www.cs.cmu.edu/afs/cs/project/PIE/MultiPie.}) and two symmetrical side views (from 6 different baselines $\{b_S = {\pm} 15^\circ, b_M = {\pm} 30^\circ, b_L = {\pm} 45^\circ$, $b_{60} = {\pm} 60^\circ, b_{75} = {\pm} 75^\circ, b_{90} = {\pm} 90^\circ\}$) for training and evaluation.
	Furthermore, the cropped image triplets include not only the facial region but also the complex background.

	\noindent\textbf{KITTI.}
	The outdoor dataset KITTI provides 22 odometry and image sequences of urban city scenes.
	Our experimental settings on KITTI are the same to the standard setting of view synthesis and have been widely used in view synthesis methods~\cite{zhou2016view,Flynn2016,sun2018multi,habtegebrial2018fast}.
    The KITTI~\cite{Geiger2012} dataset contains the frame sequences captured by the camera on a car traveling through urban city scenes.
    When we select two frames from a sequence, they can be seen as two views captured by the cameras at different positions. Actually, it is consistent with the standard setting of view synthesis that we seek to produce a novel view from multiple views captured by the cameras at different positions.
	To demonstrate the robustness and superiority of \SexyName\ under the scenes with complex backgrounds, we conduct experiments on this dataset.
	According to the settings of VSAF \cite{zhou2016view} on KITTI, 
	we first randomly sample a frame as ground truth and then select two symmetric frames within the sequence that are separated by $\pm K$ frames as the input image pair,
	where $k$ is randomly sampled from the set $\{ 1,2,...,7 \}$. 
	We split the first 11 sequences into 9 for training and 2 for testing and randomly collect paired frames with different baselines.
	
	\subsection{Comparing Methods}
	On Multi-PIE and KITTI,
	we compare \SexyName\ with several state-of-the-art methods, including View Synthesis by Appearance Flow (VSAF)~\cite{zhou2016view}
	, Deep View Morphing (DVM)~\cite{Ding2017}, View Independent Generative Adversarial Network  (VI-GAN)~\cite{Xu_2019_ICCV}, Multi-Scale Adversarial Correlation Matching (MS-ACM)~\cite{zhang2019structure}, and Extreme View Synthesis (EVS)~\cite{choi2019extreme}.
	Since the source code of DVM, MS-ACM and VI-GAN are not public, we reimplement DVM and MS-ACM on TensorFlow~\cite{abadi2016tensorflow} and VI-GAN on PyTorch~\cite{NEURIPS2019_9015}.
	
	We also consider the widely used image-content-based method Encoder-Decoder (ED)~\cite{Ding2017} in the comparisons. ED is inspired by the Encoder-Decoder Network (EDN) in DVM. We employ the ED as a baseline to evaluate the content-based method. To this end, we modify the outputs and architecture of the decoders in the EDN to have the same input and output settings as our method.
	We also extend the adversarial training scheme to DVM~\cite{Ding2017} and obtain its GAN-based variant, called adversarial DVM (ADVM). 
	
	Furthermore, to demonstrate the excellent performance of our method, we further compare \SexyName\ with some generative methods, \eg, Pix2Pix~\cite{isola2017image} and CycleGAN~\cite{zhu2017unpaired}, and geometry-based methods that leverage known camera poses or plane-sweep volumes, \eg, Multi2Novel~\cite{sun2018multi} and StereoMagnification~\cite{zhou2018stereo}, on KITTI.  
	
	\begin{figure*}[ht]
		\centering
		\includegraphics[width=7.3in]{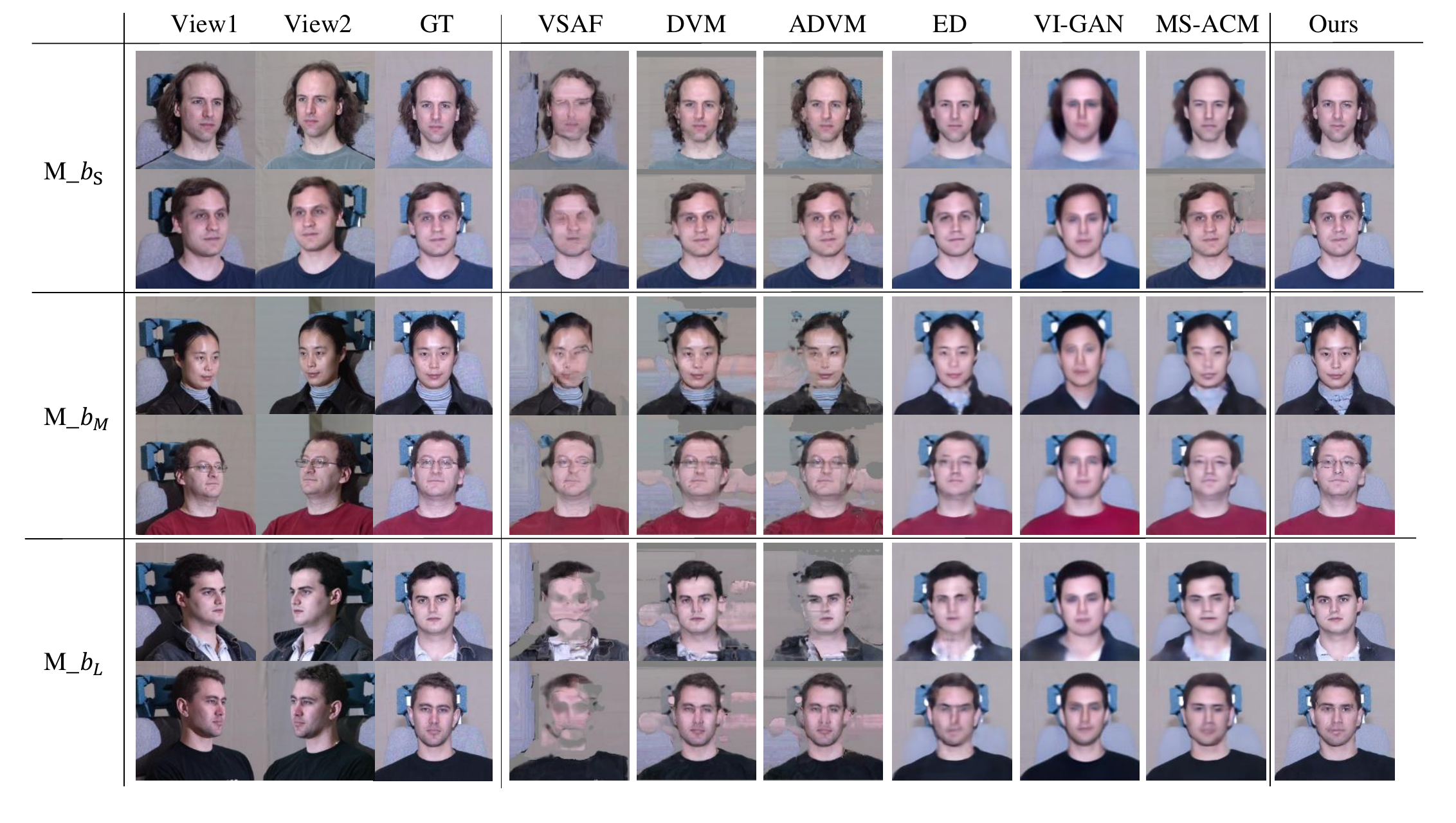}
		\vspace{-1cm}
		\caption{Visual comparisons of the synthesized view images under different baselines, where $M$ represents Multi-PIE and $\{b_S, b_M, b_L\}$ are three different baselines. For Multi-PIE, $\{$View1, View2, GT$\}$ represent $\{$left, right, middle$\}$ views. }
		\label{fig:comp_all}
		\vspace{-0.3cm}
	\end{figure*}
	
	\subsection{Evaluation Metrics}
	\label{sec:evalMetrics}
	
	For quantitative comparisons, we adopt the \emph{Peak Signal-to-Noise Ratio} (PSNR), \emph{Multi-Scale Structural Similarity} (MS-SSIM)~\cite{karras2017progressive}, and \emph{Inception Score} (IS)~\cite{salimans2016improved} as performance metrics.
	The PSNR measures the amount of signal loss \wrt a reference and the MS-SSIM measures the similarity between the generated images and the reference images.
	The inception score measures both the single image quality and the diversity over a large number of samples. 
	For all the above metrics, the larger the metric value is, the better the performance of the method is.
	We also adopt \emph{the mean of MSE} (mMSE) (used in DVM~\cite{Ding2017}) and \emph{$L_1$ error} (used in VSAF \cite{zhou2016view}) for fair comparisons. 
	For these two metrics, the smaller the metric value is, the better the performance of the method is.
	
	Furthermore, several view synthesis quality assessment methods~\cite{dist2019, zhou2019no, huang2020blind, li2020quality} have been shown very effective for view synthesis quality evaluation.
	Specifically, we compare the performances of our SCGN model with the considered methods in terms of LOGS~\cite{quality2018}, which is a view synthesis quality metric. For the LOGS, the higher metric value indicates better quality.

	\begin{figure}[t]
		\centering
		\includegraphics[width=3.3in]{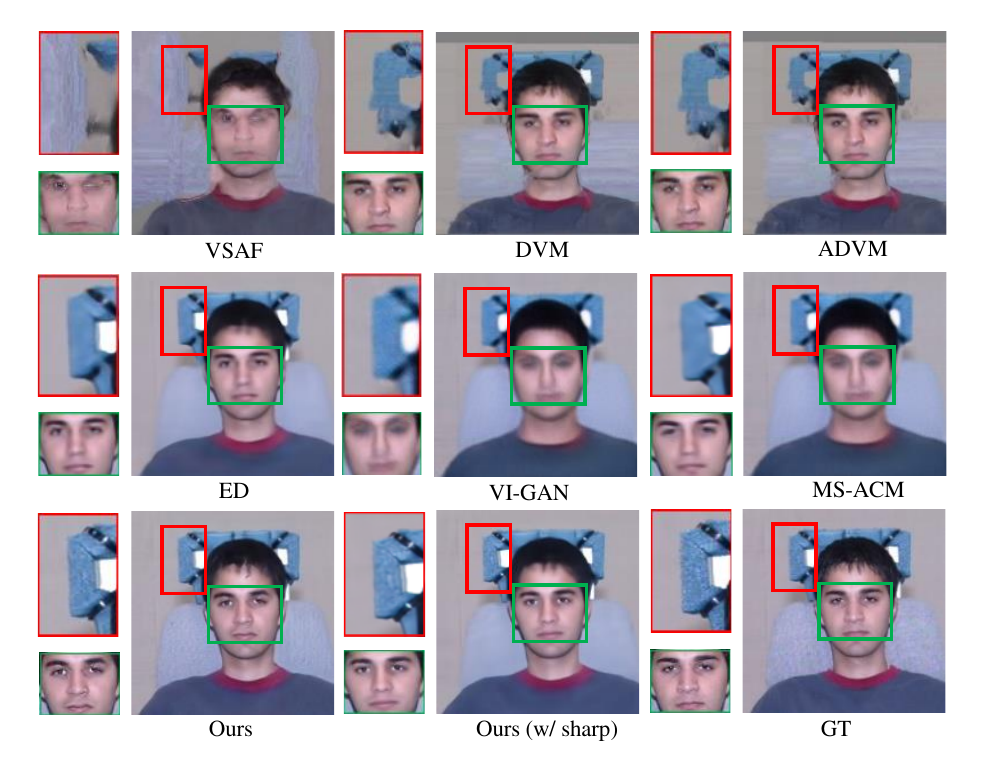}
		\vspace{-0.4cm}
		\caption{Visual comparison of image details generated by different methods on Multi-PIE. The red boxes and the green boxes emphasize the local details of background and face, respectively.}
		\label{fig:comp_issue_bg}
		\vspace{-0.4cm}
	\end{figure}
	
	\subsection{Training Convergence}
	\label{exp:convergence}
	
	In this experiment, we compare the training and testing convergence of different methods on both Multi-PIE and KITTI datasets in terms of PSNR and MS-SSIM. 
	The experimental results of the indoor and outdoor datasets are shown in Fig.~\ref{fig:psnr_msssim} and Fig.~\ref{fig:psnr_msssim_kitti}, where (a), (b) and (c), (d) show the convergence results in terms of PSNR and MS-SSIM, respectively, during both training and testing periods. 
	
	From Fig.~\ref{fig:psnr_msssim} and Fig.~\ref{fig:psnr_msssim_kitti}, our \SexyName\ shows faster convergence than other methods in terms of both PSNR and MS-SSIM. 
	As for the testing performance, our \SexyName\ consistently outperforms other methods during the whole training procedure. 
	These results demonstrate the superior performance of the proposed method over competing approaches.

	\begin{figure*}[th]
    \centering
    \includegraphics[width=7in]{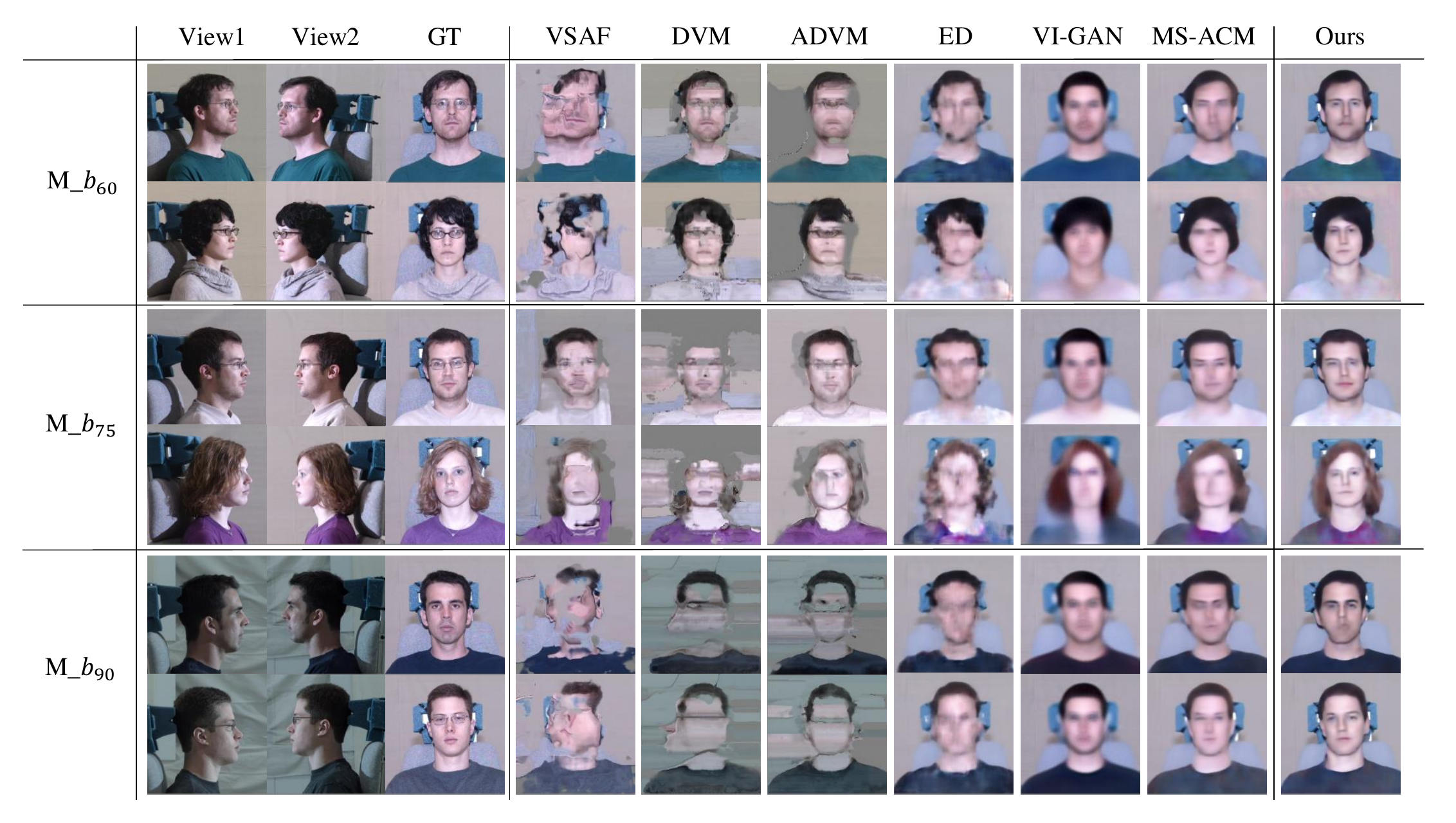}
    \vspace{-0.2cm}
    \caption{Visual comparisons of the synthesized view images under different baselines, where $M$ represents Multi-PIE and $\{b_{60}, b_{75}, b_{90}\}$ are three large different baselines. For Multi-PIE, $\{$View1, View2, GT$\}$ represent $\{$left, right, middle$\}$ views. }
    \label{fig:respone_sota_multipie_large}
    \vspace{-0.4cm}
\end{figure*}
	
	\subsection{View Synthesis Results on Indoor Scenes}
	\label{sec:indoor}

	We compare our \SexyName\ with state-of-the-arts on Multi-PIE and report the results in Table~\ref{tab:error_small}. Noted that the results on Multi-PIE for EVS are unavailable because of the lack of camera pose in the Multi-PIE.
	From Table~\ref{tab:error_small}, we observe that, first, 
	\SexyName\ consistently outperforms other methods on all evaluation metrics under different baseline settings. 
	Our model obtains the highest PNSR and inception score, suggesting that the synthesized results are more photo-realistic.
	Moreover, high MS-SSIM values and high LOGS values with low error values (\eg, mMSE, \emph{$L_1$ error}) show that \SexyName\ generates accurate results.
	Second, \SexyName\ significantly outperforms the considered methods, especially when the baseline is very large, \ie, $b_M$, $b_L$, $b_{60}$, $b_{75}$, and $b_{90}$. 
	In other words, our method is able to effectively predict the frontal view from two distant side views. 
	In contrast, geometry-based methods (\eg, VSAF, DVM, ADVM, EVS) obtain poor metric values and fail to synthesize high-quality frontal views under large baselines. 
	Furthermore, ED performs better than these geometry-based methods, which shows that view synthesis under large baselines benefits from methods based on image content.

	From the visual comparison results in Fig.~\ref{fig:comp_all}, VSAF and DVM struggle to produce plausible frontal views when increasing the baseline from small to large.
	For the modified version ADVM with an additional adversarial loss, the image quality still suffers due to the limitations of 2D geometric approximation.
	The Encoder-Decoder (ED) exhibits good performance when given a small baseline but fails to produce plausible images under a large baseline. 
	In contrast, \SexyName\ is able to recover photo-realistic frontal views when given different baselines.
	We also apply our method to the view synthesis tasks on the Multi-PIE dataset with three large baselines, \emph{i.e.}, $b_{60}$, $b_{75}$, and $b_{90}$. We show the results in Fig.~\ref{fig:respone_sota_multipie_large} and Table~\ref{tab:error_large}.
    From these results, our \SexyName~significantly outperforms the considered methods. From Fig.~\ref{fig:vdn_outputs}, our method effectively preserves the consistency among different views.
	
	To show the effectiveness of our method for synthesizing photo-realistic frontal views, we exhibit the detailed structure and texture of the results produced by different methods in Fig.~\ref{fig:comp_issue_bg}.
	Clearly, \SexyName\ is able to produce quality images with sharper face structures and finer details in the background.

	\begin{table}[t]
		\caption{Performance comparison with state-of-the-art view synthesis methods under three baselines $b_S, b_M, b_L$ on Multi-PIE.}
		\small
		\begin{center}
			\begin{minipage}[b]{1\linewidth}
				\centering
				\resizebox{1.0\textwidth}{!}{
					\begin{tabular}{c|c|c|c|c|c|c}
						\hline
						\multicolumn{7}{c}{\textbf{Baseline} $\mathbf{b_S}$}\\
						\hline
						Method & PSNR & MS-SSIM & Inception Score & mMSE & $L_1$ & LOGS\\
						\hline
						VSAF~\cite{zhou2016view} & 19.56 &0.7199 &-& 134.50 & 0.171& 0.1979 \\
						DVM~\cite{Ding2017} & 21.26& 0.8686&-&131.67&0.123&0.2581 \\	
						ADVM &20.89& 0.8587& 1.68$\pm$0.12&131.76&0.125 &0.2544 \\
						ED~\cite{Ding2017} & 23.30 &0.8940 &-&130.63&0.083 &0.2505 \\
                     	VI-GAN~\cite{Xu_2019_ICCV} & 21.22  & 0.7910 & 1.72$\pm$0.19  & 131.87 & 0.101& 0.2436  \\
				 	    MS-ACM~\cite{zhang2019structure} & 25.16  & 0.9096 & 1.62$\pm$0.38 & 129.23 & 0.064 &0.2501 \\
						\SexyName\ (ours) & \textbf{26.36}&\textbf{0.9620}&\textbf{1.97$\pm$0.22}&\textbf{128.95}&\textbf{0.054}& \textbf{0.2719} \\
						\hline
					\end{tabular}
				}
			\end{minipage}
			\\
			\begin{minipage}[b]{1\linewidth}
				\centering
				\resizebox{1.0\textwidth}{!}{
					\begin{tabular}{c|c|c|c|c|c|c}
						\hline
						\multicolumn{7}{c}{\textbf{Baseline} $\mathbf{b_M}$}\\
						\hline
						Method & PSNR & MS-SSIM & Inception Score & mMSE & $L_1$& LOGS\\
						\hline
						VSAF~\cite{zhou2016view} & 17.35 &0.6273 &-&136.39&0.191& 0.1799 \\
						DVM~\cite{Ding2017} & 18.39& 0.6744& -&135.52&0.180& 0.2069\\	
						ADVM & 18.25& 0.7010& 1.67$\pm$0.16&135.37&0.185&0.1746 \\
						ED~\cite{Ding2017} & 21.07 &0.8270 &-&132.54&0.103 &0.2278 \\
						VI-GAN~\cite{Xu_2019_ICCV} & 20.35  & 0.7565 & 1.72$\pm$0.17 & 132.82 & 0.111& 0.2416 \\
				        MS-ACM~\cite{zhang2019structure} & 22.51 & 0.8464 & 1.66$\pm$0.41 & 130.71 & 0.086 & 0.2182 \\
						\SexyName\ (ours) & \textbf{22.83}& \textbf{0.8578} &\textbf{1.97$\pm$0.22}& \textbf{130.40} & \textbf{0.076} & \textbf{0.2459}\\
						\hline
					\end{tabular}
				}
			\end{minipage}
			\\
			\begin{minipage}[b]{1\linewidth}
				\centering
				\resizebox{1.0\textwidth}{!}{
					\begin{tabular}{c|c|c|c|c|c|c}
						\hline
						\multicolumn{7}{c}{\textbf{Baseline} $\mathbf{b_L}$}\\
						\hline
						Method & PSNR & MS-SSIM & Inception Score & mMSE & $L_1$& LOGS\\
						\hline
						VSAF~\cite{zhou2016view} & 16.27 &0.5257 &-&139.01&0.215& 0.1860 \\
						DVM~\cite{Ding2017} & 16.77& 0.6431& -&139.12&0.224& 0.2144\\
						ADVM & 17.09& 0.6436& 1.67$\pm$0.22&138.94&0.223& 0.2034\\
						ED~\cite{Ding2017} & 20.51 &0.7741 &- &133.30&0.112& 0.2229\\
						VI-GAN~\cite{Xu_2019_ICCV} & 19.66   & 0.7277  & 1.77$\pm$0.27 &133.92  & 0.121& 0.2346 \\
				        MS-ACM~\cite{zhang2019structure} &  21.57 & 0.8174 & 1.66$\pm$0.39  & 131.50  & 0.096& 0.2177 \\
						\SexyName\ (ours) & \textbf{21.75} & \textbf{0.8268} & \textbf{1.91$\pm$0.24} & \textbf{131.37} &\textbf{0.087}& \textbf{0.2407}\\
						\hline
					\end{tabular}
				}
			\end{minipage}
		\end{center}
		\label{tab:error_small}
		\vspace{-0.5cm}
	\end{table}

	\begin{table}[t]
		\caption{Performance comparison with state-of-the-art view synthesis methods under three baselines $b_{60},b_{75},b_{90}$ on Multi-PIE.}
		\small
		\begin{center}
			\begin{minipage}[b]{1\linewidth}
			\centering
			\resizebox{1.0\textwidth}{!}{
				\begin{tabular}{c|c|c|c|c|c|c}
					\hline
					\multicolumn{7}{c}{\textbf{Baseline} $\mathbf{b_{60}}$}\\
					\hline
					Method & PSNR & MS-SSIM & Inception Score & mMSE & $L_1$& LOGS\\
					\hline
					VSAF~\cite{zhou2016view} & 10.18&0.0599&-&172.78&0.405 & 0.2005\\
					DVM~\cite{Ding2017} & 18.47&0.6985&-&135.67&0.169 & 0.2220\\
					ADVM  & 16.33& 0.5990 & 1.66$\pm$0.22 &140.58 & 0.245 & 0.1620\\
					ED~\cite{Ding2017}  & 19.73	&0.7362&- &134.50&0.126 & 0.2107\\
					VI-GAN~\cite{Xu_2019_ICCV}  & 19.25 & 0.6973 & 1.71$\pm$0.23 & 134.41 & 0.130 & 0.1965\\
                    MS-ACM~\cite{zhang2019structure}  & \textbf{20.85}  &0.7780  &1.52$\pm$0.20 &\textbf{132.29} &  0.104 & 0.1997 \\
					\SexyName\ (ours) & 20.84 &\textbf{0.7929}  & \textbf{1.84$\pm$0.33}&132.32 &  \textbf{0.102} & \textbf{0.2443}  \\
						\hline
				\end{tabular}
			}
		\end{minipage}
		    \\
	    	\begin{minipage}[b]{1\linewidth}
			\centering
			\resizebox{1\textwidth}{!}{
				\begin{tabular}{c|c|c|c|c|c|c}
					\hline
					\multicolumn{7}{c}{\textbf{Baseline} $\mathbf{b_{75}}$}\\
					\hline
					Method & PSNR & MS-SSIM & Inception Score & mMSE & $L_1$& LOGS \\
					\hline
					VSAF~\cite{zhou2016view} &15.97&0.5563&-&139.38&0.201  & 0.1763\\
					DVM~\cite{Ding2017}& 15.48&0.4513&-&143.32&0.267 & 0.1817  \\
					ADVM & 17.17& 0.5876&	1.78$\pm$0.21 & 138.50 & 0.182 & 0.2191\\
					ED~\cite{Ding2017} & 18.22 & 0.6602 & - &  137.32 & 0.152 & 0.2058 \\
					VI-GAN~\cite{Xu_2019_ICCV} &18.57  &0.6666  &\textbf{1.80$\pm$0.30} & 135.62& 0.140 & 0.1962\\
                    MS-ACM~\cite{zhang2019structure}   &\textbf{19.45}  & 0.7331 & 1.60$\pm$0.24 & \textbf{134.09}&  0.122 & 0.2030 \\
					\SexyName\ (ours) &19.41 & \textbf{0.7459} & 1.71$\pm$0.24 & 134.11 & \textbf{0.121} & \textbf{0.2317} \\
					\hline
				\end{tabular}
			}
		\end{minipage}
		    \\
		    \begin{minipage}[b]{1\linewidth}
			\centering
			\resizebox{1\textwidth}{!}{
				\begin{tabular}{c|c|c|c|c|c|c}
					\hline
					\multicolumn{7}{c}{\textbf{Baseline} $\mathbf{b_{90}}$}\\
					\hline
					Method & PSNR & MS-SSIM & Inception Score & mMSE & $L_1$ & LOGS\\
			        \hline
					VSAF~\cite{zhou2016view}  &8.99&0.1165&-&187.32&0.549 & 0.2156\\
					DVM~\cite{Ding2017} & 14.61 & 0.4409&	-	& 146.66& 0.316 & 0.1994\\
					ADVM  & 14.92 & 0.4825 & \textbf{2.36$\pm$0.36} & 145.47& 0.297 & 0.2140 \\
					ED~\cite{Ding2017}  & 17.68	& 0.6528 & - &	138.43 &	0.160 & 0.2075\\
					VI-GAN~\cite{Xu_2019_ICCV}  & 18.53 & 0.6782 & 1.88$\pm$0.38 & 136.13& 0.142 & 0.1972\\
                    MS-ACM~\cite{zhang2019structure}  & 19.27 & 0.7282 &1.65$\pm$0.27 &134.85 & 0.127 & 0.1840 \\
					\SexyName\ (ours) & \textbf{19.29 }& \textbf{0.7418} & 1.74$\pm$0.28 & \textbf{134.65} & \textbf{0.123} & \textbf{0.2258}  \\
					\hline
				\end{tabular}
			}
		\end{minipage}
		\end{center}
		\label{tab:error_large}
		\vspace{-0.5cm}
	\end{table}

	\begin{figure*}[ht]
		\centering
		\vspace{0.4cm}
		\includegraphics[width=7.1in]{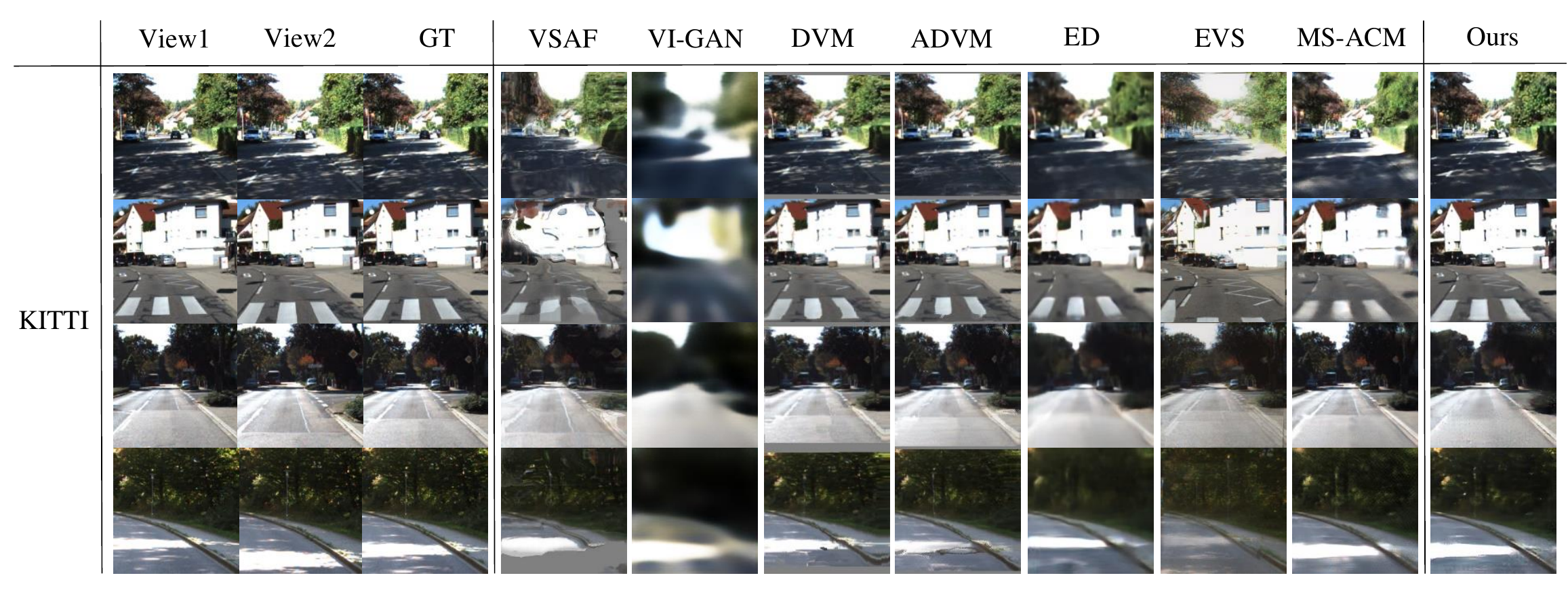}
		\vspace{-0.7cm}
		\caption{Visual comparisons of the synthesized view images on KITTI, where $\{$View1, View2, GT$\}$ represent $\{$last, next, current$\}$ frames.}
		\label{fig:comp_kitti}
	\end{figure*}
	
	\subsection{View Synthesis Results on Outdoor Scenes}
	\label{sec:outdoor}
	
	Quantitative comparisons using the evaluation metrics are shown in Table~\ref{tab:error_kitti} and Table~\ref{tab:error_kitti_add}.	
	From Table~\ref{tab:error_kitti}, \SexyName\ outdistances the state-of-the-art view synthesis methods on PSNR and MS-SSIM, although the inception score of ADVM is slightly higher than that of our method.
	From Table~\ref{tab:error_kitti_add}, our method consistently outperforms the other methods according to several metrics, which verifies its effectiveness.
	It is worth noting that the image-to-image translation task is different from view synthesis because the latter has to synthesize a novel view from two distinct views with more strict constraints (\eg, generation of occluded contents), and it is hard to handle the synthesis task under large baselines for the compared geometry-based methods in Table~\ref{tab:error_kitti_add} which work well under stereo settings or other small baselines.
	
	\begin{figure*}[t]
		\centering
		\includegraphics[width=1\linewidth]{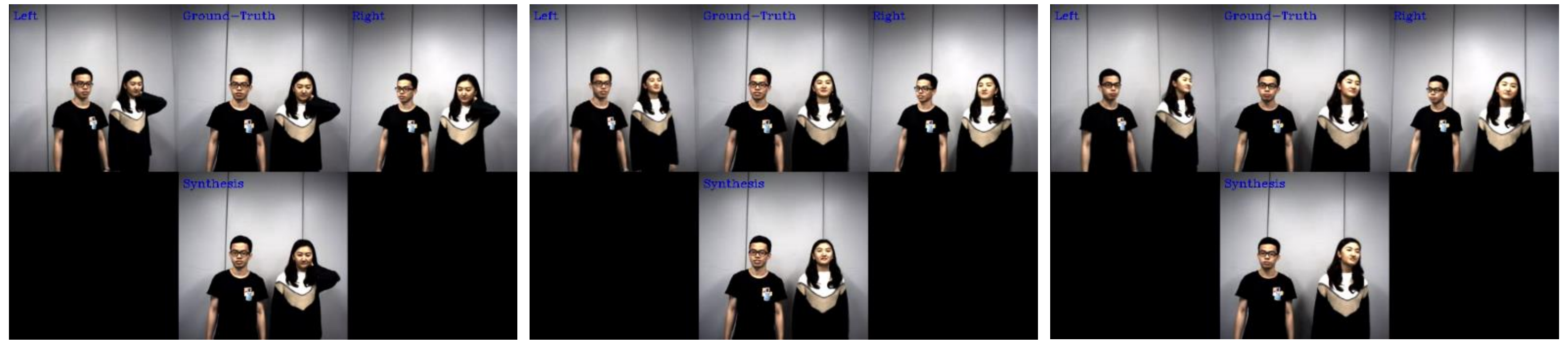}
		\vspace{-0.2in}
		\caption{Visual results of our demo in a real-world video conferencing system with a pre-defined baseline between the left and right cameras.}
		\label{fig:demo_good_results}
		\vspace{-0.2in}
	\end{figure*}
	
\begin{figure}[t]
		\centering
		\includegraphics[width=3.5in]{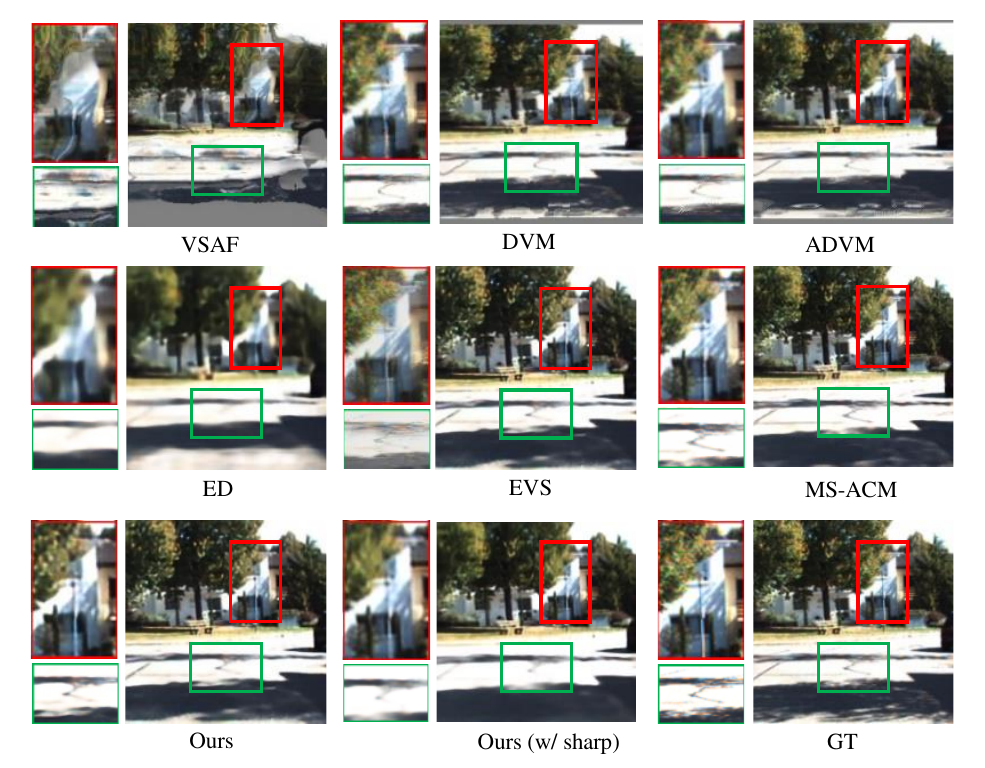}
		\vspace{-0.3in}	
		\caption{Visual comparisons of image details generated by different methods on KITTI. The red boxes and the green boxes emphasize the local details of tree texture and shadow on the road, respectively.}
		\label{fig:comp_issue_bg_kitti}
		\vspace{-0.4cm}
	\end{figure}

	We show the experimental results of visual comparison in Fig.~\ref{fig:comp_kitti}. 
	For VSAF and the Encoder-Decoder (ED), the synthesized frontal views contain many deformations and blurs. 
	For DVM and ADVM, artifacts still appear in some regions even through the generated images look realistic on the whole.  
	In contrast, \SexyName\ maintains robustness and performs well in scenes with rich texture, complex background and different light conditions.
	We also show image details such as shadows on the road in Fig.~\ref{fig:comp_issue_bg_kitti}, which further demonstrates the superiority of our methods in terms of details. 
	
	\vspace{-0.3cm}
	\begin{table}[t]
		\vspace{0.3cm}
		\caption{Performance comparison with state-of-the-art view synthesis methods under different baselines on KITTI.}
		\begin{center}
			\begin{minipage}[b]{1\linewidth}
				\centering
				\resizebox{1.0\textwidth}{!}{
					\begin{tabular}{c|c|c|c|c|c|c}
						\hline
						Method & PSNR & MS-SSIM &Inception Score & mMSE &$L_1$ & LOGS \\
						\hline
						VSAF~\cite{zhou2016view} & 13.87 &0.4533 &-&151.34&0.258 & 0.3417\\
						VI-GAN~\cite{Xu_2019_ICCV} & 15.05  & 0.5294  &1.83$\pm$ 0.06  &146.31  & 0.233&0.1239 \\
						DVM~\cite{Ding2017} & 15.48& 0.6552& -&144.25&0.205 & 0.4051\\
						ADVM & 16.30& 0.6861& 2.96$\pm$0.29&141.37&0.179& 0.3750  \\
						ED~\cite{Ding2017} & 17.28 &0.6859&-&139.28&0.159& 0.3365 \\
						EVS~\cite{choi2019extreme} & 14.74  & 0.5135 & - &148.93  & 0.250&0.3776\\
				        MS-ACM~\cite{zhang2019structure}  & \textbf{19.35} & 0.7715  & \textbf{3.78$\pm$0.50} & 135.37 & 0.124 & 0.3817 \\
						\SexyName\ (ours) & 19.20 & \textbf{0.7772}& 2.41$\pm$0.25&\textbf{129.48}&\textbf{0.031}&  \textbf{0.4097}\\
						\hline
					\end{tabular}
				}
			\end{minipage}
		\end{center}
		\label{tab:error_kitti}
	\end{table}

	\begin{table}[!tp]
		\vspace{0.18cm} 
		\caption{Comparisons with additional state-of-the-art methods on KITTI. All methods are trained and tested in a paired setup.}
		\label{tab:error_kitti_add}
		\begin{center}
			\begin{minipage}[b]{1\linewidth}
				\centering
				\resizebox{0.8\textwidth}{!}{
					\begin{tabular}{c|c|c|c}
						\hline
						Method & PSNR & MS-SSIM & $L_1$ \\
						\hline
						Pix2Pix~\cite{isola2017image} & 12.59 & 0.6943 & 0.141 \\
						CycleGAN~\cite{zhu2017unpaired} & 8.17 & 0.5112 & 0.226 \\
						Multi2Novel~\cite{sun2018multi} & 10.36 & 0.3904 & 0.413 \\
						StereoMagnification~\cite{zhou2018stereo} & 11.87 & 0.3792 & 0.194 \\
						\SexyName\ (ours) & \textbf{19.20} & \textbf{0.7772} & \textbf{0.031}  \\
						\hline
						
					\end{tabular}
				}
			\end{minipage}
		\end{center}
	\end{table}

	\begin{table}[t]
		\small
		\caption{Ablation study of each model component on KITTI. We compare the results in terms of PSNR, MS-SSIM, and Inception Score where $m$ represents the modified version.}
		\vspace{-0.2cm}
		\begin{center}
			
			\begin{tabular}{c|c|c|c|c|c}
				\hline
				VSN & VDN &  adv & PSNR & MS-SSIM &Inception Score \\
				\hline
				$\checkmark$ &    &   & 18.96 & 0.7382 & - \\
				$m$ &  &  & 18.92 & 0.7351  & - \\
				$\checkmark$ &    & $\checkmark$  & 18.80& 0.7533&\textbf{2.45$\pm$0.22} \\
				$\checkmark$  &  $m$   & $\checkmark$ & 18.81 & 0.7432  & 2.38$\pm$0.20 \\
				$\checkmark$  &  $\checkmark$  &  & 19.16& 0.7511& - \\
				$\checkmark$   &  $\checkmark$   & $\checkmark$ & \textbf{19.20} & \textbf{0.7773} & 2.41$\pm$0.25 \\
				\hline
			\end{tabular}
		\end{center}
		\label{tab:without}
		\vspace{-0.3cm}
	\end{table}	
	
	\begin{table}[!t]
		\caption{Ablation results (PSNR) of VDN on Multi-PIE.}
		\label{tab:without_vdn_multipie}
		\vspace{-0.2cm}
		\centering
		\resizebox{0.4\textwidth}{!}{
			\begin{tabular}{c|c|c|c|c|c}
				\hline
				VSN & VDN &  adv & $\mathbf{b_S}$ & $\mathbf{b_M}$ & $\mathbf{b_L}$ \\
				\hline
				$\checkmark$ &    & $\checkmark$ & 25.52 & 22.59 & 21.54 \\
				$\checkmark$  &  $m$   & $\checkmark$ & 25.59 & 22.69 & 21.68 \\
				$\checkmark$   &  $\checkmark$   & $\checkmark$ & \textbf{26.36} & \textbf{22.83} & \textbf{21.75} \\
				\hline
			\end{tabular}
		}
	    \vspace{-0.1in}
	\end{table}	
	
	\subsection{Results on Real-world Conferencing System}	
	
	\subsubsection{Details about the conferencing dataset}
	\label{sec:demo}
	For demo evaluation, we set up the experimental environment by placing two cameras on the left side and the right side, and capture view pairs as our inputs. 
	Furthermore, to capture the frontal view as the ground-truth, we place an additional camera at the center of the screen on the same horizontal line as the other cameras.
	
	\subsubsection{Implementation details}
	We train the \SexyName\ model on our conferencing dataset with a pre-trained model which is trained on Multi-PIE. 
	Note that the pre-trained data are with similar settings of our conferencing dataset, \textit{i.e.}, people with a background in symmetrical image pairs.
	We also record two conferencing demos to further demonstrate the effectiveness and robustness of our \SexyName. One of the demos shows the real conferencing system with less actions (named as \emph{\SexyName\_demo\_talk.mp4}), and the other shows a scene with much more rapid movements (named as \emph{\SexyName\_demo\_move.mp4}).
	
	\subsubsection{View synthesis results on our demos}
	As shown in Fig.~\ref{fig:demo_good_results}, for most scenes, our \SexyName\ performs effectively and synthesizes views as photo-realistic as the captured frontal view. The good-quality visual results of the demos that even including two people with occlusions demonstrate the robustness of our SCGN. From the talking demo with fewer actions, we observe that the synthesized views are excellent with the small changes of inputs, and more details and visualized results of the moving demo are shown in the supplementary.

	
	\section{Further Experiments}
	\label{sec:ablation}
	
	We conduct further experiments on KITTI and Multi-PIE to demonstrate the effectiveness of each component of \SexyName, including the View Synthesis Network (VSN), the View Decomposition Network (VDN), the adversarial loss, and the sharpness loss.

	\subsection{Effect of View Synthesis Network}
	\label{sec:effectVSN}
	
	We investigate the effect of VSN by comparing the original version of VSN (w/o VDN \& adv) and the modified version of VSN (mVSN). 
	Compared to VSN, mVSN removes the max-pooling layers of the encoder and the upsampling layers of the decoder.
	In addition, we train these two versions of VSN using only the $L_p$ loss in Eq.~(\ref{eq:loss_g}).
	As shown in Table \ref{tab:without}, all of the evaluation metrics show that the original version of VSN outperforms the mVSN, which demonstrates the necessity of the feature compression and extraction mechanism in VSN.
	
	 \begin{figure*}[thb]
    	\centering
    	\includegraphics[width=7in]{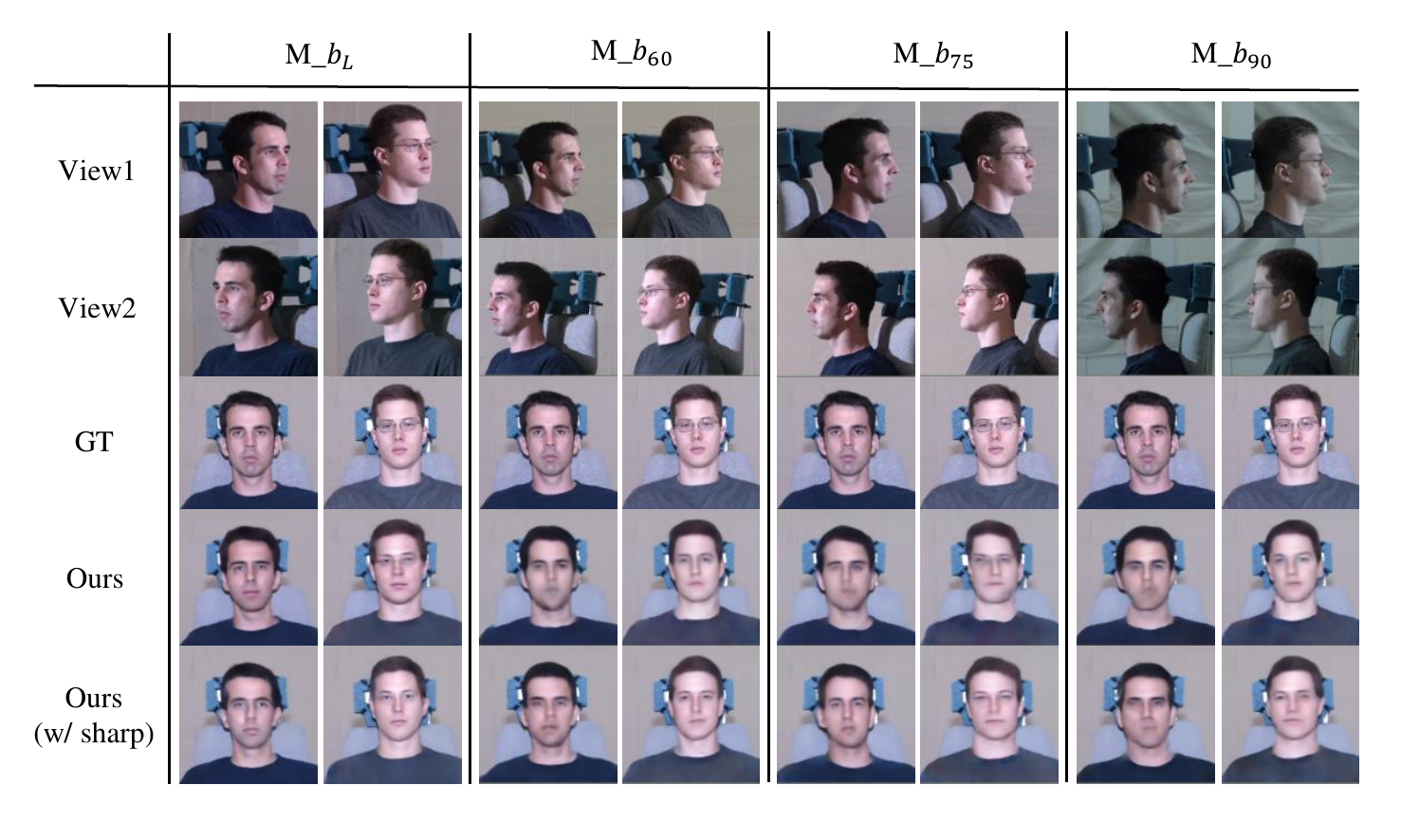}
    	\vspace{-0.4cm}
    	\caption{Visual comparisons of our \SexyName~and a variant with the sharpness loss $L_{sharp}$ (w/ sharp) under different baselines, where $M$ represents Multi-PIE and $\{b_L, b_{60}, b_{75}, b_{90}\}$ are four different baselines in Multi-PIE..}
    	\label{fig:sharp_large_multipie}
    	\vspace{-0.4cm}
    \end{figure*}  
    
    	\begin{figure}[ht]
		\centering 
		\includegraphics[width=3in]{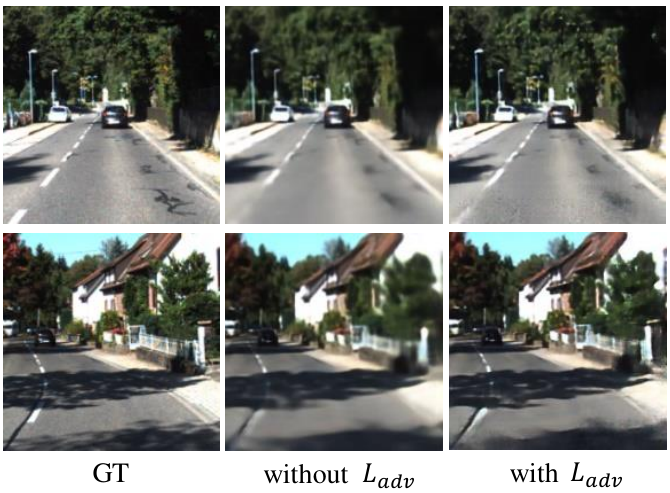}
		\caption{Performance comparisons of \SexyName\ with and without the adversarial loss $L_{adv}$ on KITTI dataset.}
		\label{fig:adv_comp}
		\vspace{-0.2in}
	\end{figure}
	
	\subsection{Effect of View Decomposition Network}
	\label{sec:effectVDN}
	
	We investigate the effect of VDN by removing the VDN component (w/o VDN) and removing the $L_{vc}$ in Eq.~(\ref{eq:loss_g}).
	Furthermore, we modify the separated decoders $\{V^{D, l}, V^{D, r}\}$ to be a single weight-shared decoder (mVDN) to evaluate the effect of the separated decoders in VDN.
	
	As shown in Table \ref{tab:without}, our model with VDN (\SexyName) significantly outperforms the model without VDN on KITTI in terms of both PSNR and MS-SSIM although the inception score of our model is slightly lower than that without VDN. Our model with VDN (\SexyName) also has higher PSNR than that without VDN on different baselines of Multi-PIE as shown in Table~\ref{tab:without_vdn_multipie}.
	Our model (\SexyName) also outperforms the mVDN in terms of all evaluation metrics. These results demonstrate that using the two separate decoders to obtain the decomposed side views is more effective than using a single decoder in VDN.
	
\begin{figure}[t]
		\centering
		\includegraphics[width=3.4in]{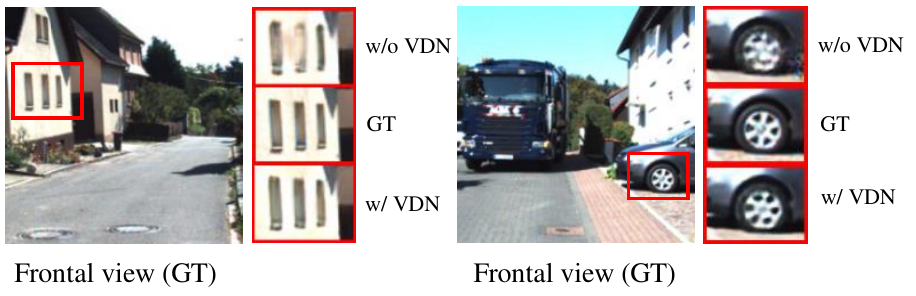}
		\vspace{-0.3cm}
		\caption{Performance comparisons of \SexyName\ with (w/) and without (w/o) View Decomposition Network (VDN). The red boxes emphasize the local details of windows and wheels, respectively.}
		\label{fig:vdn_comp}
	\end{figure}

\begin{figure}[!tp]
		\centering
		\includegraphics[width=3.3in]{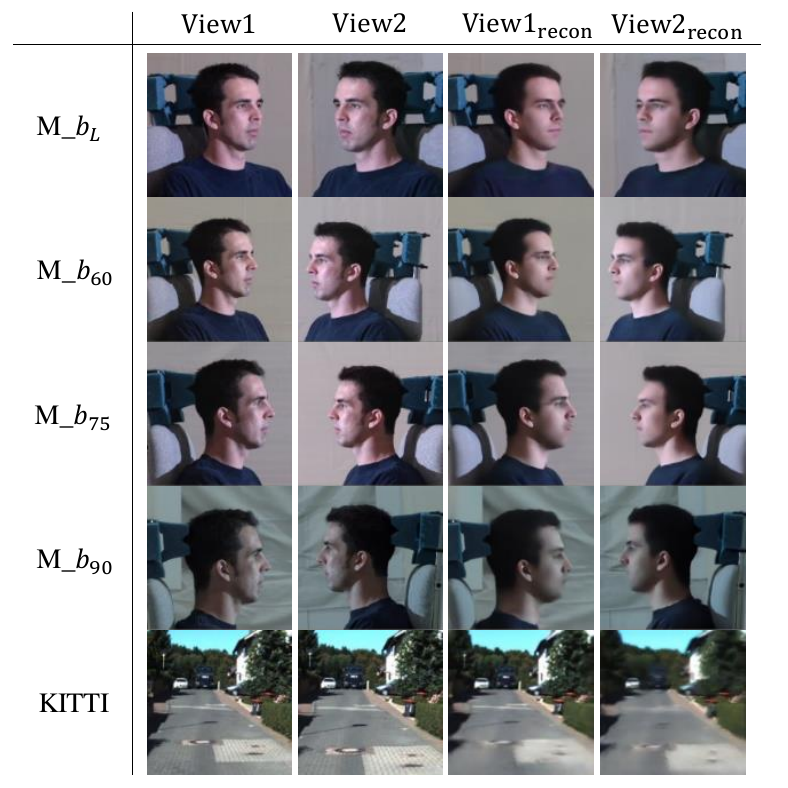}
		\caption{Regenerated results of VDN on Multi-PIE with large baselines, \emph{i.e.}, $b_L$ ($\pm45^\circ$), $b_{60}$ ($\pm60^\circ$), $b_{75}$ ($\pm75^\circ$), and $b_{90}$ ($\pm90^\circ$) and on KITTI.}
		\label{fig:vdn_outputs}
		\vspace{-0.15in}
	\end{figure}

	We also show the images regenerated by VDN in Fig.~\ref{fig:vdn_outputs}. From this figure, VDN is able to decompose the synthesized view into the original side views and recover the content details on both Multi-PIE and KITTI. Moreover, from Fig. \ref{fig:vdn_comp}, compared to the synthesized results from a variant of \SexyName\ without VDN, the windows and wheels synthesized by \SexyName\ with VDN are more realistic than those of the ground-truth (GT) with less deformation.
	All of the above comparisons show that VDN can help resolve the correspondence matching issue and compensate for the lack of rectification, so that VSN does not need to perform any geometric processing in advance, and can directly learn the photo-realistic synthesized views based on image content.

      \begin{figure*}[thb]
    	\centering
    	\includegraphics[width=7.2in]{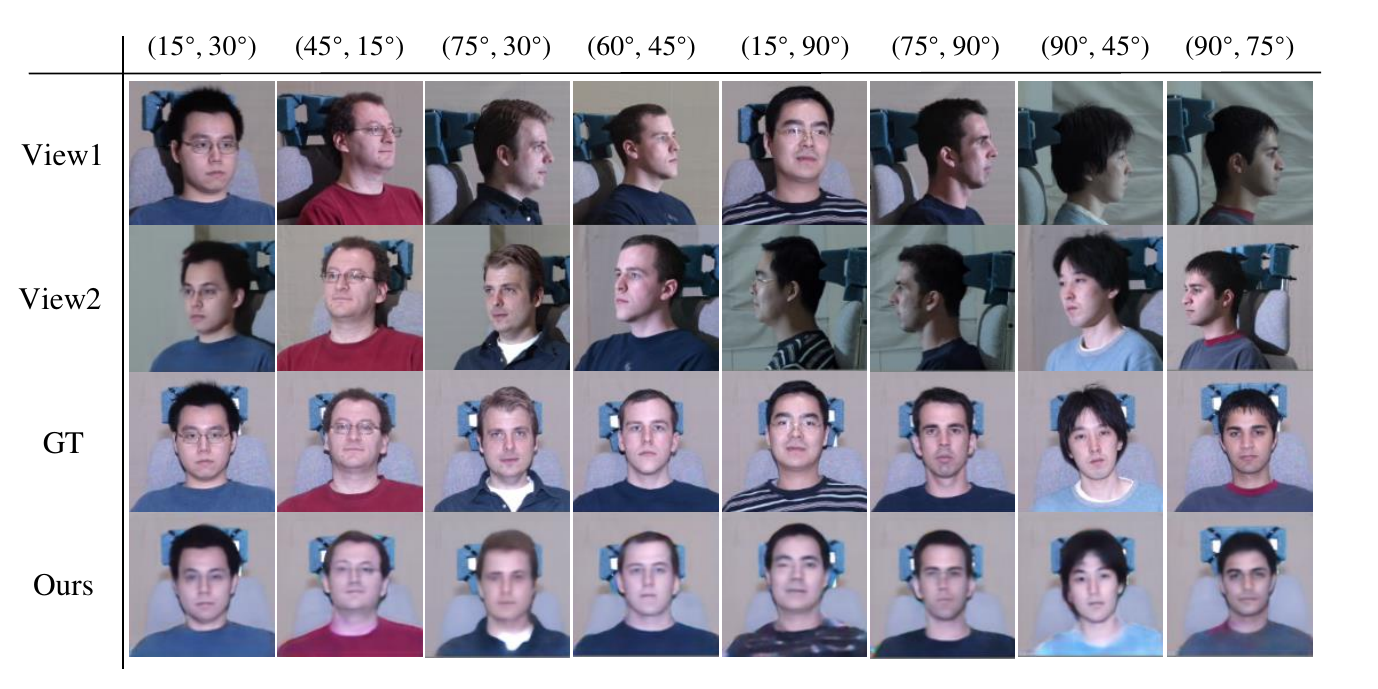}
    	\vspace{-0.35in}
    	\caption{Visual comparisons of the synthesized view under asymmetric input views, where $(\cdot,\cdot)$ represents input baseline of View1 and View2.}
    	\label{fig:response_asym}
    	\vspace{-0.4cm}
    \end{figure*} 
    
    \begin{table}[ht]
    \centering
    \caption{Ablation Study of sharpness loss $L_{sharp}$ on large baselines $b_L$, $b_{60}$, $b_{75}$, $b_{90}$ in Multi-PIE. $b_L$, $b_{60}$, $b_{75}$, $b_{90}$ indicate baseline $\pm45^\circ$, $\pm60^\circ$, $\pm75^\circ$, $\pm90^\circ$ respectively. }
    \vspace{-0.05in}
	\resizebox{1\linewidth}{!}{
    \begin{tabular}{c|c|c|c|c|c|c}
        \hline
     	\multicolumn{7}{c}{\textbf{Baseline} $\mathbf{b_{L}}$}\\
         \hline
         $L_{sharp}$ & PSNR & MS-SSIM & Inception Score & mMSE & $L_1$ & LOGS  \\
          \hline
          \xmark & 21.75 & 0.8268 & \textbf{1.91$\pm$0.24} & 137.37 & \textbf{0.087} & 0.2329 \\
         \checkmark & \textbf{22.03} & \textbf{0.8349} & 1.83$\pm$0.30 & \textbf{131.10} & \textbf{0.087} & \textbf{0.2391} \\
        \hline
        \hline
        \multicolumn{7}{c}{\textbf{Baseline} $\mathbf{b_{60}}$}\\
        \hline
       $L_{sharp}$ & PSNR & MS-SSIM & Inception Score & mMSE & $L_1$ & LOGS  \\
        \hline
         \xmark & 20.84 & 0.7929 & \textbf{1.84$\pm$0.33} & 132.32 & 0.102 & \textbf{0.2443}  \\
        \checkmark & \textbf{20.93} & \textbf{0.7969} & 1.78$\pm$0.32 & \textbf{132.23} & \textbf{0.101} & 0.2383\\
        \hline
        \hline
         \multicolumn{7}{c}{\textbf{Baseline} $\mathbf{b_{75}}$}\\
        \hline
        $L_{sharp}$ & PSNR & MS-SSIM & Inception Score & mMSE & $L_1$ & LOGS  \\
        \hline
         \xmark & 19.41 & 0.7459 & 1.71$\pm$0.24 & 134.11 & 0.121 & 0.2317 \\
        \checkmark & \textbf{19.55} & \textbf{0.7473} & \textbf{1.75$\pm$0.31} & \textbf{133.90} & \textbf{0.117} & \textbf{0.2374}\\
        \hline
        \hline
         \multicolumn{7}{c}{\textbf{Baseline} $\mathbf{b_{90}}$}\\
        \hline
        $L_{sharp}$ & PSNR & MS-SSIM & Inception Score & mMSE & $L_1$ & LOGS  \\
        \hline
        \xmark & 19.29 & 0.7418 & \textbf{1.74$\pm$0.28} & 134.65 & 0.123 & 0.2258  \\
        \checkmark & \textbf{19.42} & \textbf{0.7458} & 1.68$\pm$0.25 & \textbf{134.42} & \textbf{0.120} & \textbf{0.2342} \\
        \hline
    \end{tabular}
    }
    \vspace{-0.15in}
    \label{tab:mpie_sharpness}
\end{table}

\subsection{Effect of the Adversarial Loss $L_{adv}$} 
	\label{sec:effectAdv}
	We investigate the effect of adversarial learning by removing $L_{adv}$ (\ie, Eq.~(\ref{eq:loss_adv})) from the training procedure. 
	As shown in Table \ref{tab:without}, the PSNR, MS-SSIM, and inception score of \SexyName\ without adversarial loss is slightly higher than that with adversarial loss. 
	The adversarial loss introduces diversity to improve the photo-realism, leading to lower evaluation metrics. 
	From visual qualitative comparisons in Fig. \ref{fig:adv_comp}, it can be seen that \SexyName\ synthesizes rich details such as a clear lane line, tree texture, and shadow on the road, while the \SexyName\ without the adversarial loss synthesizes smooth results.
	\vspace{-0.1in}

	\begin{table}[!t]
		\begin{center}
			\caption{Comparison of the average inference latency and the performance of different methods on KITTI dataset.}
			\label{tab:runtime}
			\vspace{-0.15in}
			\resizebox{0.5\textwidth}{!}{
				\begin{tabular}{c|c|c|c|c|c}
					\hline
					Method & VSAF~\cite{zhou2016view} & DVM~\cite{Ding2017} & ADVM & ED~\cite{Ding2017} & \SexyName \\
					\hline
					Inference Latency (s) & 0.037 & 0.052 & 0.064 & 0.068 & \textbf{0.036}\\
					PSNR & 13.87 & 15.48 & 16.30 & 17.28 & \textbf{19.20}\\
					\hline
				\end{tabular}
			}
		\end{center}
		\vspace{-0.2in}
	\end{table}
	
	    \begin{table}[!t]
	    \vspace{-0.05in}
        \centering
        \caption{Ablation Study of asymmetric baseline in Multi-PIE, where Asym. represents training \SexyName~with asymmetric inputs views.}
        \begin{tabular}{c|c|c|c|c|c}
             \hline
             Asym. & PSNR & MS-SSIM & Inception Score & mMSE & $L_1$   \\
             \hline
             \xmark &  21.75  & \textbf{0.8212 } & 1.86$\pm$0.26 & 131.99 & \textbf{0.094 }  \\
             \checkmark  & \textbf{21.77} & 0.8179 & \textbf{2.14$\pm$0.06} & \textbf{131.48} & 0.095 \\
            \hline
        \end{tabular}
        \label{tab:asym_multipie}
    \end{table}
	
    \subsection{Effect of the Sharpness Loss $L_{sharp}$}
    \label{sec:sharpness}
    
    To investigate the effect of $L_{sharp}$ in Eq.~\ref{eq:sharpness}, we consider 4 view synthesis settings on large baselines, \emph{i.e.}, $b_L$ ($\pm45^\circ$), $b_{60}$ ($\pm60^\circ$), $b_{75}$ ($\pm75^\circ$), and $b_{90}$ ($\pm90^\circ$). We show the results in Table~\ref{tab:mpie_sharpness} and Fig.~\ref{fig:sharp_large_multipie}. From the results, the model with the sharpness term outperforms the baseline model in most cases.

    \subsection{Discussion on Asymmetric Input Views}
    \label{sec:asymmetric}
    We apply our method to the view synthesis tasks with asymmetric input views on the Multi-PIE dataset.
    In the experiments, we randomly sample angles from the range between $15^\circ$ to $90^\circ$ to construct the asymmetric views.
    From the results in Table~\ref{tab:asym_multipie} and Fig.~\ref{fig:response_asym}, our \SexyName~is able to produce photo-realistic views from asymmetric input views.

    \subsection{Effect of $\lambda_1$, $\lambda_2$, and $\lambda_3$ on the Performance of \SexyName}
	\label{sec:moreDiscuss}
	
	In this section, we investigate the effect of  $\lambda_1$, $\lambda_2$, and $\lambda_3$ in Eq.~(\ref{eq:loss_g}) on Multi-PIE and KITTI. Table \ref{tab:lambdasOnes} shows the experimental results with different $\lambda_1$ values when $\lambda_2 = 0.001$ and $\lambda_3=0.01$. 
	The results for $\lambda_1=0.01$ are better than the others in terms of PSNR, MS-SSIM, and Inception Score (IS) on both datasets. 
	We also evaluate our method with different $\lambda_2$ values when $\lambda_1=0.01$ and $\lambda_3=0.01$. 
	Table \ref{tab:lambdasTwo} shows that \SexyName\ with $\lambda_2=0.001$ achieves the best performance on three metrics.
	In addition, we investigate our method with different values of $\lambda_3$ when $\lambda_1=0.01$ and $\lambda_2=0.001$. From Table~\ref{tab:lambdasThree}, our method performs the best when $\lambda_3 = 0.01$ in terms of three metrics.
	As a result, we suggest setting $\lambda_1=0.01$, $\lambda_2=0.001$, and $\lambda_3=0.01$ for \SexyName\ by default.

	\begin{table}[!t]
		\small
		\caption{Effect of $\lambda_1$ on the performance of \SexyName.}\label{tab:lambdasOnes}
		\begin{center}
			\vspace{-0.2in}
			\begin{minipage}[b]{1\linewidth}
				\centering
				\resizebox{1.0\textwidth}{!}{
					\begin{tabular}{c|c|c|c|c|c|c}
						\hline
						\multirow{2}{*}{$\lambda_1$} & \multicolumn{3}{c|}{Multi-PIE} & \multicolumn{3}{c}{KITTI} \\
						\cline{2-7}
						& PSNR & MS-SSIM & IS & PSNR & MS-SSIM & IS\\
						\hline
						0.1 & 23.73 & 0.8722 & \textbf{2.34$\pm$0.30} & 18.03 & 0.6957 & 2.50$\pm$0.30\\
						0.01 & \textbf{23.95} &\textbf{0.8774} & 2.19$\pm$0.24 & \textbf{19.20}& \textbf{0.7773}& 2.41$\pm$0.25  \\
						0.001 & 23.08 & 0.8511 & 2.06$\pm$0.27  & 18.80 & 0.7433 & \textbf{2.53$\pm$0.23}\\
						\hline
					\end{tabular}
				}
			\end{minipage}
			\vspace{-0.2in}
		\end{center}
	\end{table}
	
	\begin{table}[!t]
		\small
		\caption{Effect of $\lambda_2$ on the performance of \SexyName.}\label{tab:lambdasTwo}
		\begin{center}
			\vspace{-0.2in}
			\begin{minipage}[b]{1\linewidth}
				\centering
				\resizebox{1.0\textwidth}{!}{
					\begin{tabular}{c|c|c|c|c|c|c}
						\hline
						\multirow{2}{*}{$\lambda_2$} & \multicolumn{3}{c|}{Multi-PIE} & \multicolumn{3}{c}{KITTI} \\
						\cline{2-7}
						& PSNR & MS-SSIM & IS & PSNR & MS-SSIM & IS\\
						\hline
						0.01 & 23.65 & 0.8655 & \textbf{2.24$\pm$0.30} & 18.53 & 0.7338 & 2.32$\pm$0.18\\
						0.001 & \textbf{23.95} &\textbf{0.8774} & 2.19$\pm$0.24 & \textbf{19.20}& \textbf{0.7773}& \textbf{2.41$\pm$0.25} \\
						0.0001 & 23.06 & 0.98498 & 1.99$\pm$0.22 & 18.49 & 0.7262 & 2.28$\pm$0.21  \\
						\hline
					\end{tabular}
				}
			\end{minipage}
		\end{center}
		\vspace{-0.2in}
	\end{table}
	
		\begin{table}[!t]
		\small
		\caption{Effect of $\lambda_3$ on the performance of \SexyName.}\label{tab:lambdasThree}
		\begin{center}
			\vspace{-0.2in}
			\begin{minipage}[b]{1\linewidth}
				\centering
				\resizebox{1.0\textwidth}{!}{
					\begin{tabular}{c|c|c|c|c|c|c}
						\hline
						\multirow{2}{*}{$\lambda_3$} & \multicolumn{3}{c|}{Multi-PIE} & \multicolumn{3}{c}{KITTI} \\
						\cline{2-7}
						& PSNR & MS-SSIM & IS & PSNR & MS-SSIM & IS\\
						\hline
						0.1 & 19.71 & 0.7924 & 1.80$\pm$0.22 & 17.02 & 0.6840 & 2.52$\pm$0.32\\
						0.01 & \textbf{23.95} &\textbf{0.8774} & \textbf{2.19$\pm$0.24} & \textbf{19.86} & \textbf{0.7706} & \textbf{3.56$\pm$0.41}  \\
						0.001 & 23.08 & 0.8511 & 2.06$\pm$0.27 & 19.73 & 0.7605 & 3.53$\pm$0.53 \\
						\hline
					\end{tabular}
				}
			\end{minipage}
			\vspace{-0.2in}
		\end{center}
	\end{table}
	
	\subsection{Comparison of the Inference Latency of Different Methods}
	\label{sec:runtime}
	
	In this section, we show the average inference latency of different methods on the KITTI dataset using a single Nvidia TitanX GPU. We show the comparison results of latency and performance in Table~\ref{tab:runtime}. From these results,
	our method exhibits the fastest inference speed (27 fps) but yields the best performance above all the other compared methods.
	
	
	\section{Conclusion}
	\label{sec:conclusion} 
	We have presented a simple but effective view synthesis network to synthesize unseen frontal and middle views from two side views with a large camera baseline without geometric processing.
	Specifically, we propose a view decomposition network by learning an inverse mapping from the synthesized view back to the input view pair to preserve content consistency; this mapping can take the place of rectification and solve the pixel-level matching problem. 
	To improve the photo-realism of images, we further introduce an adversarial loss to increase the likelihood that the synthesized images will be indistinguishable from the real views. 
	As a result, the proposed method can simultaneously produce photo-realistic unseen views and preserve the view consistency among all views of the same scene. 
	Using different baselines, the proposed method consistently outperforms the other methods in terms of both quantitative and visual comparisons.

	\section*{Acknowledgements}
	This work was partially supported by the Science and Technology Program of Guangzhou, China, under Grant 202007030007, the Key-Area Research and Development Program of Guangdong Province (2018B010107001), National Natural Science Foundation of China (NSFC) 61836003 (key project), Guangdong Project 2017ZT07X183, Fundamental Research Funds for the Central Universities D2191240.
	

	\bibliographystyle{IEEEtran}
	
	\vspace{-1cm}
	\begin{IEEEbiography}
		[{\includegraphics[width=1in,height=1.25in,clip,keepaspectratio]{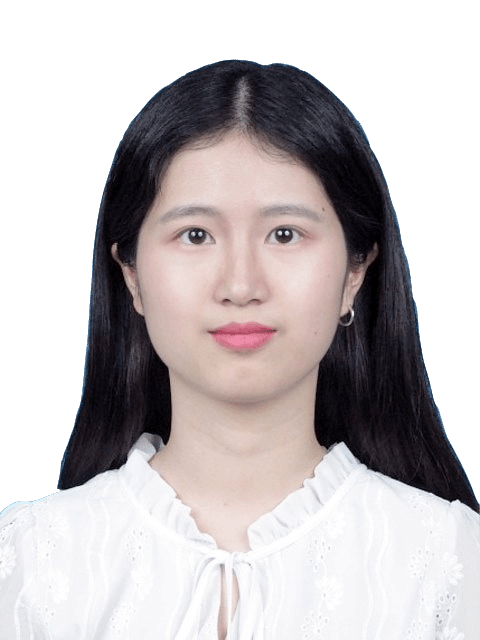}}]
		{Zhuoman Liu}
		is currently a researcher with Guangzhou Shiyuan Electronic Technology Co., Ltd (CVTE) in Guangzhou, China.
		She received the bachelor degree in Software Engineering from South China University of Technology in 2019.
		Her main research interests include deep learning and computer vision.
	\end{IEEEbiography}
	    \vspace{-1cm}
	\begin{IEEEbiography}
		[{\includegraphics[width=1in,height=1.25in,clip,keepaspectratio]{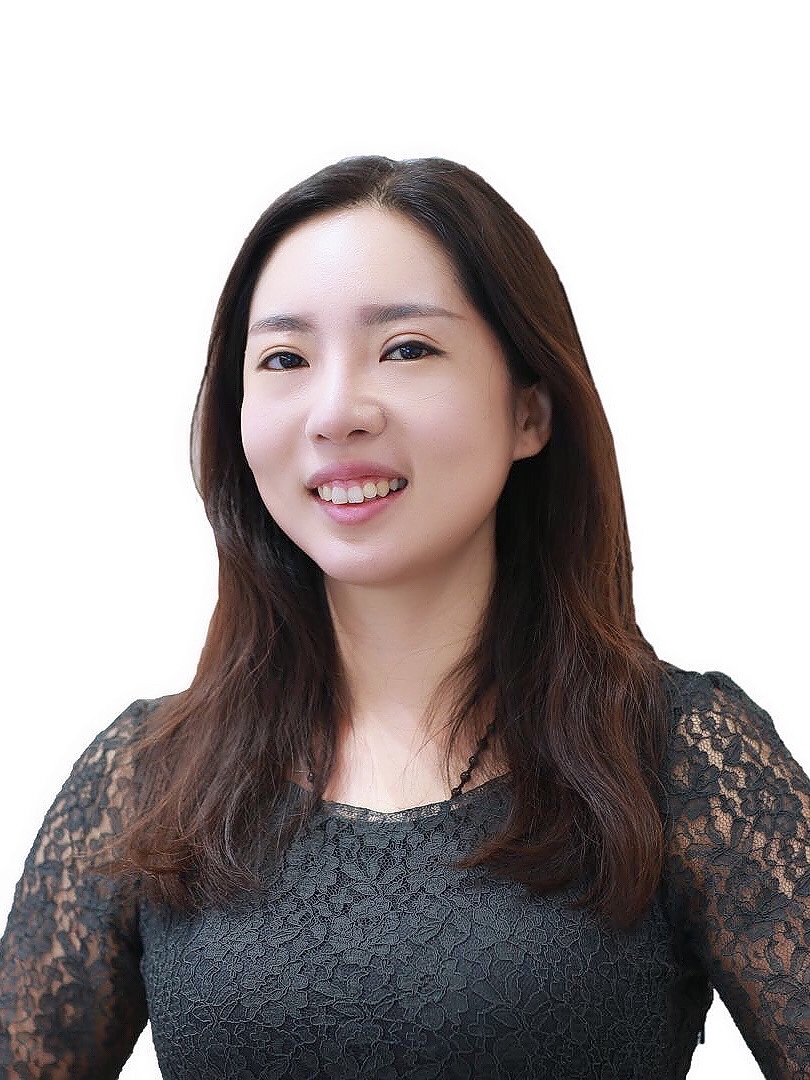}}]
		{Wei Jia}
		is currently a senior researcher at Guangzhou Shiyuan Electronic Technology Co., Ltd (CVTE) in Guangzhou, China. 
		She leads a 3D vision group and her research interests include 3D imaging, 3D reconstruction, neural networks, and deep learning. 
		She received her Ph.D degree in Computer Science from Dundee university, UK, in 2012. 
		She received her Master degree in Computer Science from University of Bristol, UK, in 2006. 
		She received her bachelor degree in Computer Science from Harbin Engineering University, China, in 2004. 
	\end{IEEEbiography}
	    \vspace{-1cm}
	\begin{IEEEbiography}
		[{\includegraphics[width=1in,height=1.25in,clip,keepaspectratio]{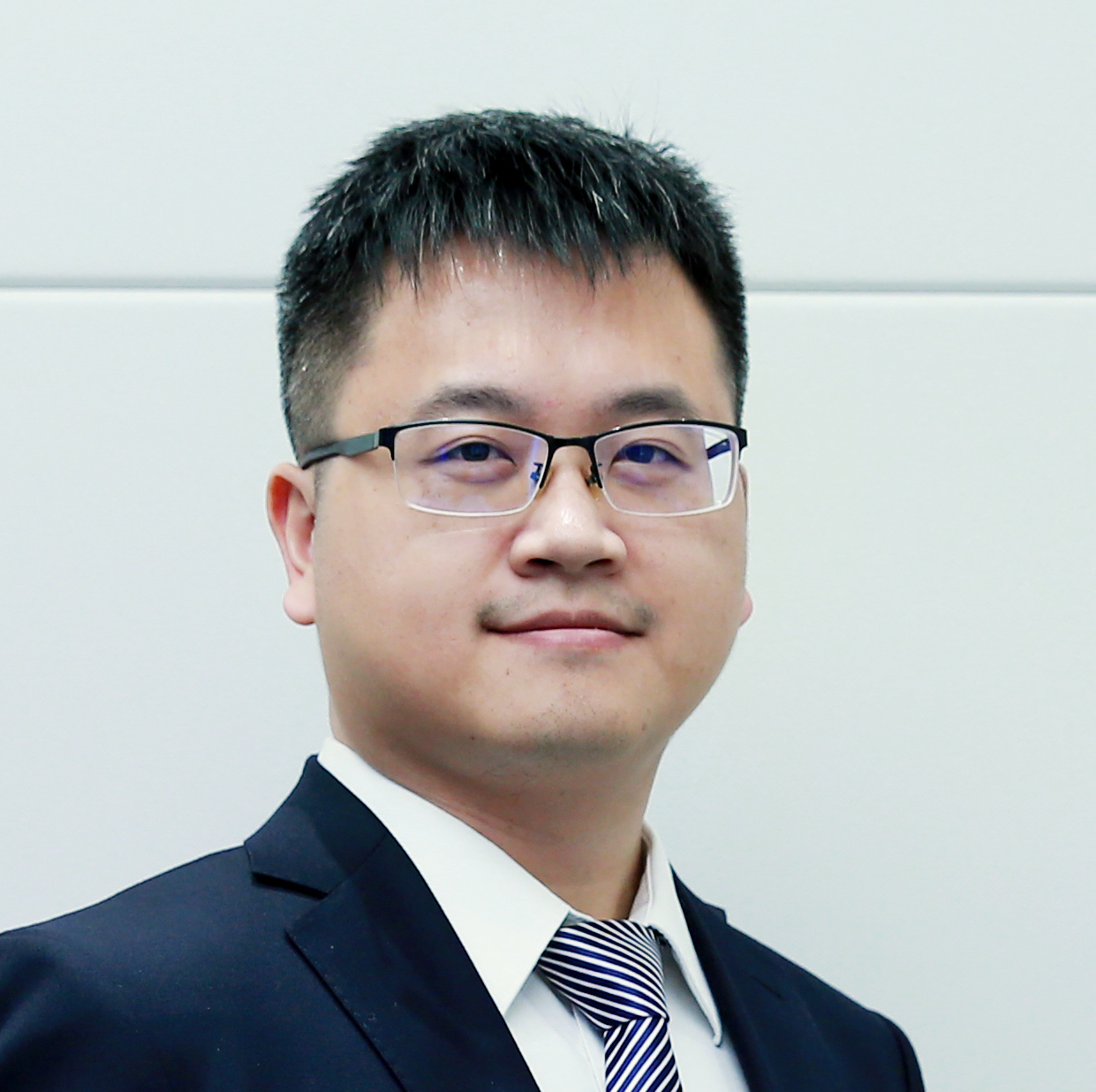}}]
		{Ming Yang}
		is currently the CTO of Guangzhou Shiyuan Electronic Technology Co., Ltd (CVTE) in Guangzhou, China. 
		He received his B.S. and Ph.D. degree from Sun Yat-sen University in 2009 and 2014, respectively. 
		He joined CVTE Research in 2014. 
		His research interests include machine learning and interactive computer vision.
	\end{IEEEbiography}
	    \vspace{-1cm}
	\begin{IEEEbiography}
		[{\includegraphics[width=1in,height=1.25in,clip,keepaspectratio]{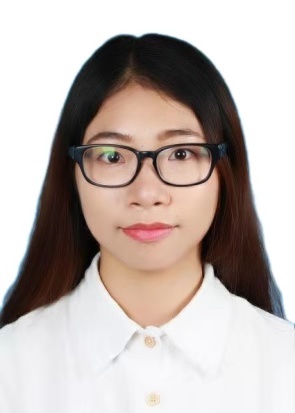}}]
		{Peiyao Luo}
		is a M.S. candidate with the School of Software Engineering at South China University of Technology.
		She also received her bachelor degree in Mechatronic Engineering from the same university in 2018.
		Her research interests include deep learning and computer vision.
	\end{IEEEbiography}
	    \vspace{-1cm}
	\begin{IEEEbiography}
		[{\includegraphics[width=1in,height=1.25in,clip,keepaspectratio]{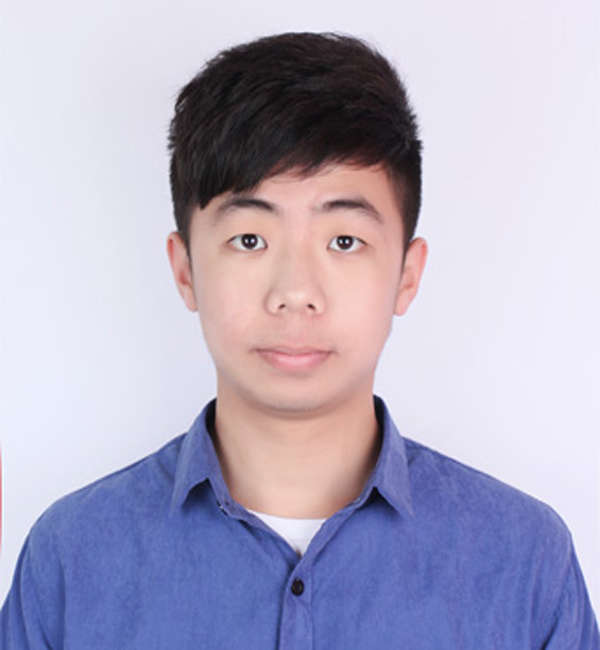}}]
		{Yong Guo}
		is a Ph.D. candidate with the School of Software Engineering at South China University of Technology.
		He also received his bachelor degree in Software Engineering from the same university in 2016.
		His research interests include deep learning and computer vision.
	\end{IEEEbiography}
	    \vspace{-1cm}
	
	\begin{IEEEbiography}
		[{\includegraphics[width=1in,height=1.25in,clip,keepaspectratio]{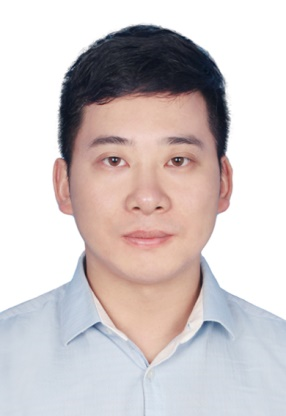}}]
		{Mingkui Tan}
		is currently a professor with the School of Software Engineering at South China University of Technology. 
		He received his Bachelor Degree in Environmental Science and Engineering in 2006 and Master degree in Control Science and Engineering in 2009, both from Hunan University in Changsha, China. 
		He received the Ph.D. degree in Computer Science from Nanyang Technological University, Singapore, in 2014. 
		From 2014-2016, he worked as a Senior Research Associate on computer vision in the School of Computer Science, University of Adelaide, Australia. 
		His research interests include machine learning, sparse analysis, deep learning and large-scale optimization.
	\end{IEEEbiography}

\end{document}


\onecolumn
	
	\title{Supplementary Materials: Deep View Synthesis \\
		via Self-Consistent Generative Network}

	\markboth{Journal of \LaTeX\ Class Files,~Vol.~14, No.~26, January~2021}%
	{Shell \MakeLowercase{\textit{et al.}}: Bare Demo of IEEEtran.cls for IEEE Journals}
	
	\maketitle
	
	
	{
	We organize our supplementary materials as follows. We first display more visual results on both Multi-PIE and KITTI datasets in Sec.~\ref{appdx:visual}. Then, we provide details of our demo in Sec.~\ref{appdx:demo}, including experimental setup and visual results on various scenes that people have large movement.
	}

	\section{More visual results on Multi-PIE and KITTI}
	\label{appdx:visual}
\begin{figure*}[!htbp]
		\centering
		\includegraphics[width=7in]{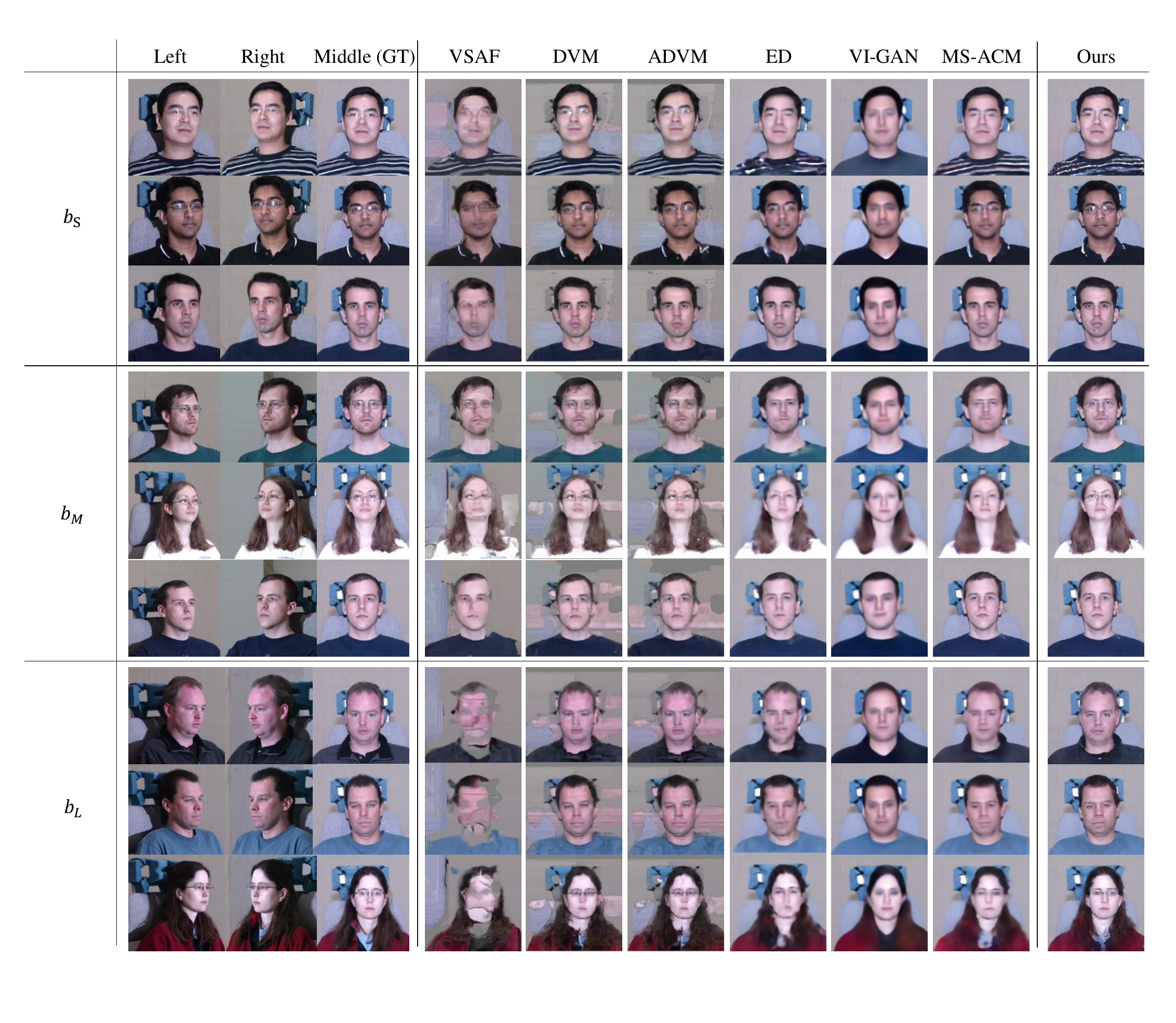}
		\caption{Visual comparisons of the synthesized middle view images under different baselines on Multi-PIE and $\{b_S, b_M, b_L\}$ are three different baselines.}
		\label{fig:comp_all_1}
	\end{figure*}

\begin{figure*}[!htbp]
		\centering
		\vspace{-2cm}
		\includegraphics[width=7in]{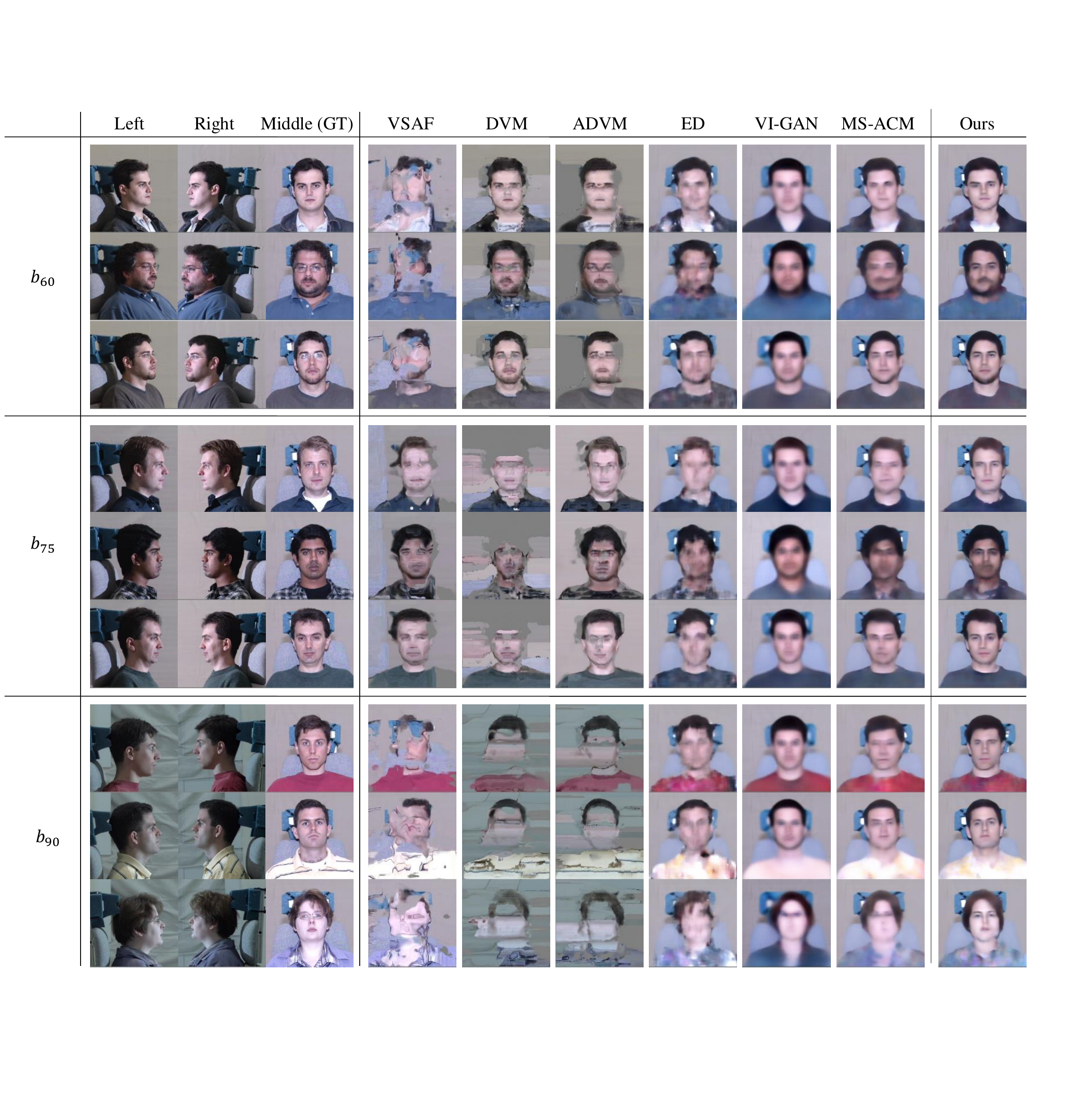}
		\caption{Visual comparisons of the synthesized middle view images under different baselines on Multi-PIE and $\{b_{60}, b_{75}, b_{90}\}$ are three different baselines.}
		\label{fig:comp_all_3}
	\end{figure*}

\begin{figure*}[!htbp]
		\centering
		\includegraphics[width=6.8in]{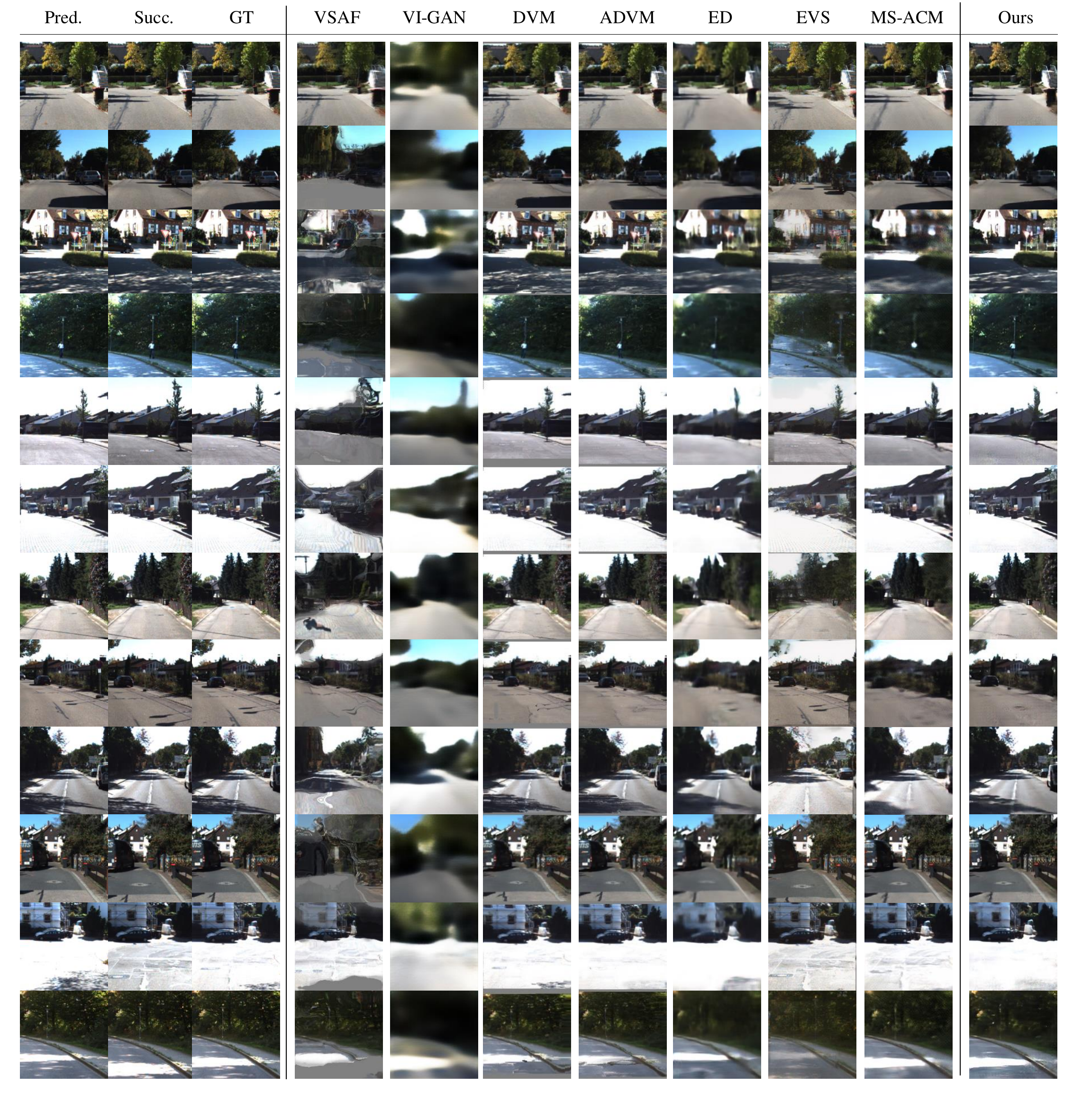}
		\caption{Visual comparisons of the synthesized middle view images under different frame intervals on KITTI.}
		\label{fig:comp_all_2}
	\end{figure*}

	We show more results on the visual comparisons between our method and other state-of-the-art methods on Multi-PIE and KITTI in Fig. \ref{fig:comp_all_1},  Fig. \ref{fig:comp_all_3} and Fig. \ref{fig:comp_all_2}, respectively. 
	From Fig. \ref{fig:comp_all_1} and Fig. \ref{fig:comp_all_3}, we observe that our method is able to synthesize detailed and photo-realistic views in both facial and background areas.
	As shown in Fig.~\ref{fig:comp_all_2}, our method can synthesize great details of outdoor scenes despite the complex contents on KITTI.

	\clearpage
	\section{Details of Our Demo}
	\label{appdx:demo}
\begin{figure}[h]
		\centering
		\includegraphics[width=2in]{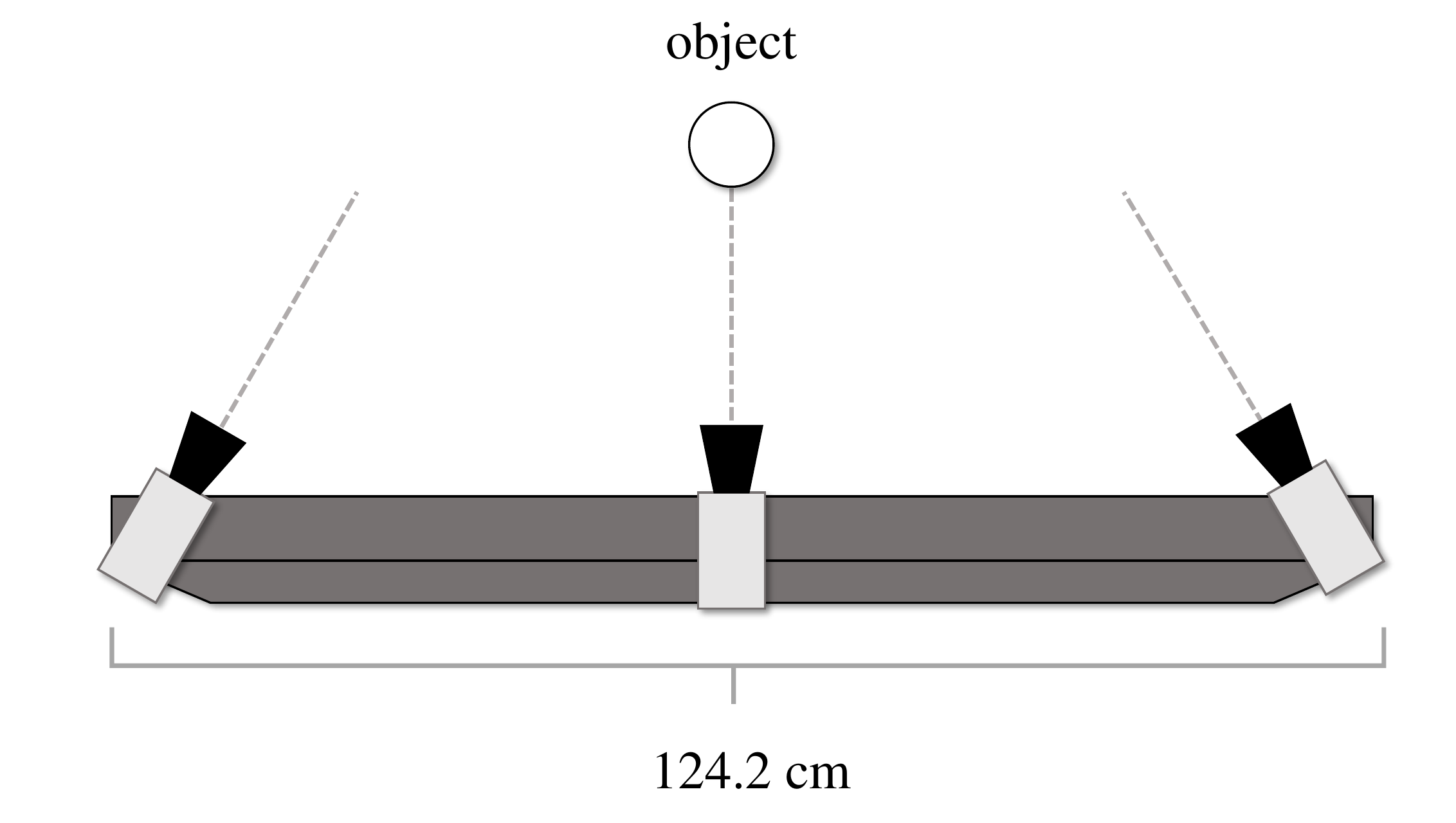}
		\caption{The setup of our conferencing system.}
		\label{fig:equipment}
	\end{figure}

{	
	We introduce the equipment that collects our conferencing dataset and demos in Fig.~\ref{fig:equipment}. Three cameras (\ie, two side cameras for capturing input views and the center camera for capturing the ground-truth) in the equipment are placed on the same horizontal line. The distance between the symmetrically placed left and right cameras is $124.2\ cm$, and the baseline of the paired view captured by the two side cameras is $b_M (\pm 30^\circ)$. All the experimental results of our proposed method on the conferencing system are evaluated on this setup.

}

\begin{figure*}[!htbp]
		\centering
		\includegraphics[width=6in]{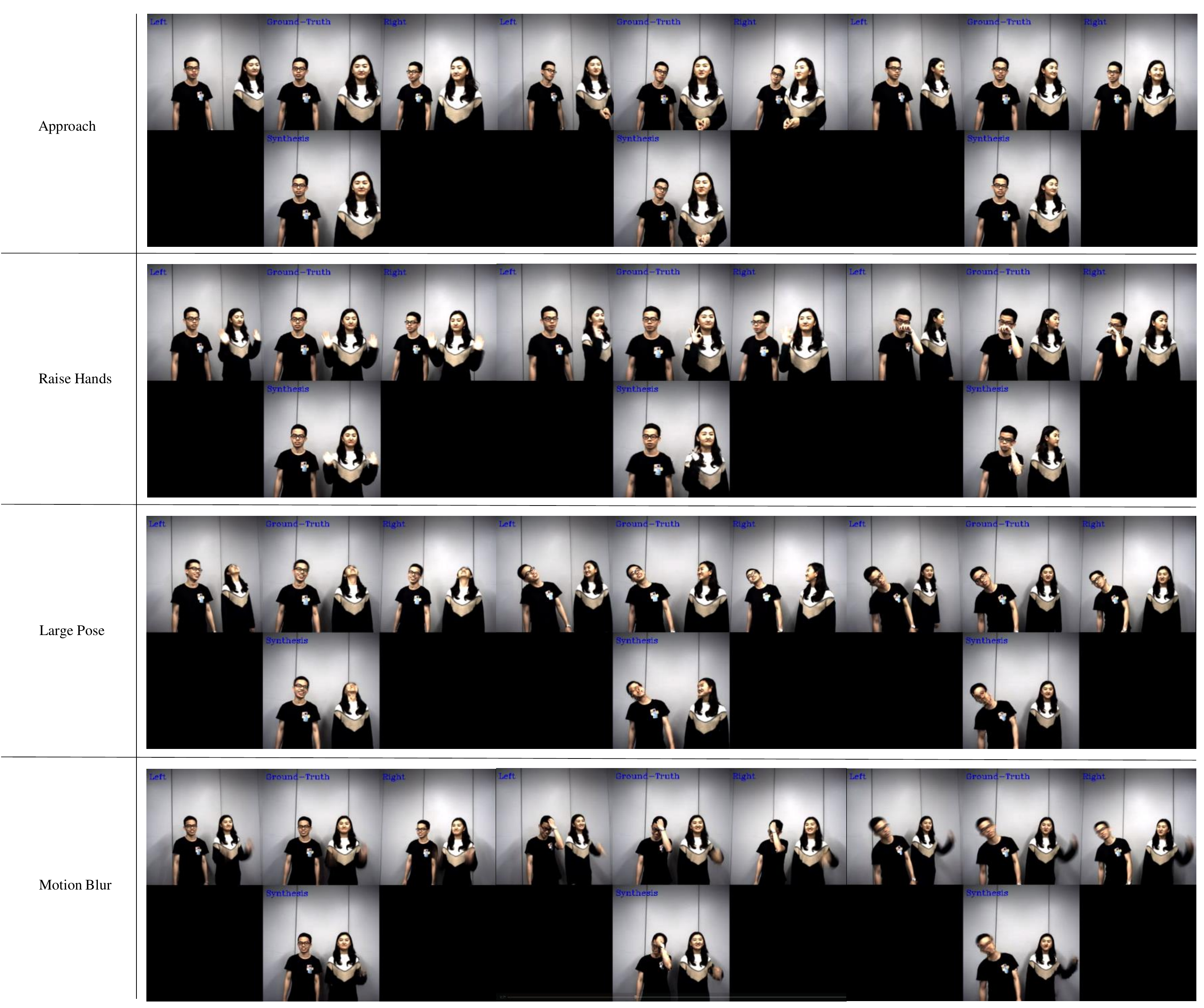}
		\caption{Details of the demo with rapid movements.}
		\label{fig:demo}
	\end{figure*}

{	

	For \emph{\SexyName\_demo\_move.mp4}, besides the good results of most scenes, we further analyze different scenes that contain much more rapid movements, and we select several frames to display in Fig.~\ref{fig:demo}. 
	\begin{itemize}
		\item \emph{Approach:}  When we approach the screen, the general synthesized views are good but distortions occur in the face area due to the non-center face position in the input view pair.
		\item \emph{Raise hands:} Due to the lack of training data with hands, our trained model is not suitable for synthesizing accurate hand areas and faces with large pose. 
		\item \emph{Large pose:} When people in front of the screen make large pose, the synthesized views may lose some details.
		\item \emph{Motion blur:} Under the situation of rapid motions which results in blurry input pairs, it is hard for our model to handle.
	\end{itemize}
	
}
